%% file: main.tex
\documentclass[runningheads]{llncs}

\usepackage[mobile]{eccv}

\usepackage{eccvabbrv}
\usepackage{graphicx}
\usepackage{booktabs}
\usepackage{colortbl}
\definecolor{Gray}{gray}{0.98}
\usepackage{bbding}
\usepackage{multirow}
\usepackage{pyhighlight}
\usepackage{algorithm}
\usepackage{tikzsymbols}
\usepackage{fontawesome}
\usepackage{tabu}
\usepackage{algorithm}
\usepackage{algorithmic}
\usepackage[accsupp]{axessibility}  
\usepackage{pyhighlight}

\usepackage[dvipsnames]{xcolor}

\usepackage{hyperref}
\usepackage{orcidlink}

\usepackage{xspace}
\def\Ours{TAP\xspace}

\title{Tokenize Anything via Prompting} 
\titlerunning{Tokenize Anything via Prompting}

\author{
Ting Pan\inst{1,2,3,}\textsuperscript{*} \and
Lulu Tang\inst{2,}\textsuperscript{*} \and
Xinlong Wang\inst{2,}\textsuperscript{\Envelope} \and
Shiguang Shan\inst{1,3}
}

\authorrunning{T~Pan et al.}

\institute{
Key Laboratory of Intelligent Information Processing, ICT, CAS \\
\email{\{ting.pan@vipl.,sgshan@\}ict.ac.cn} \and
Beijing Academy of Artificial Intelligence \\
\email{\{lltang,xlwang$\_$1\}@baai.ac.cn}  \and
University of Chinese Academy of Sciences, Beijing, China \\
\thanks{\footnotesize {*}Equal Contribution, \textsuperscript{\Envelope}Corresponding Author} 
}

\makeatletter
\def\thanks#1{\protected@xdef\@thanks{\@thanks
        \protect\footnotetext{#1}}}
\makeatother

\begin{document}

\input {arxiv_v2/insight.tex}
\maketitle
\input{arxiv_v2/0_abstract}

\input{arxiv_v2/1_intro}
\input{arxiv_v2/2_related_works}

\input{arxiv_v2/3_method}
\input{arxiv_v2/4_experiments}

\input{arxiv_v2/5_conclusion}

\clearpage
{
\small
\section*{Acknowledgement}
This project is supported by the National Key R\&D Program of China (2022ZD0116302). We would like to thank Hanxiao Qu and Yan Tian for their help on Cambricon MLU resources, as well as other colleagues at BAAI for their support to this project.
\bibliographystyle{splncs04}
\bibliography{main}
}

\clearpage
\input{arxiv_v2/arxiv_appendix}

\end{document}

%% file: arxiv_v2/0_abstract.tex
\begin{abstract}

We present a unified, promptable model capable of simultaneously segmenting, recognizing, and captioning anything. Unlike SAM, we aim to build a versatile region representation in the wild via visual prompting. To achieve this, we train a generalizable model with massive segmentation masks, \eg, SA-1B masks, and semantic priors from a pre-trained CLIP model with 5 billion parameters.
Specifically, we construct a promptable image decoder by adding a semantic token to each mask token. The semantic token is responsible for learning the semantic priors in a predefined concept space.
Through joint optimization of segmentation on mask tokens and concept prediction on semantic tokens, our model exhibits strong regional recognition and localization capabilities. 
For example, an additional 38M-parameter causal text decoder trained from scratch sets a new record with a CIDEr score of 164.7 on the Visual Genome region captioning task. 
We believe this model can be a versatile region-level image tokenizer, capable of encoding general-purpose region context for a broad range of visual perception tasks. Code and models are available at {\footnotesize \url{https://github.com/baaivision/tokenize-anything}}.

\keywords{Multi-modal \and Image Tokenization \and Regional Perception}

\end{abstract}

%% file: arxiv_v2/1_intro.tex
\section{Introduction}\label{sec:intro}

A key objective of visual perception is to efficiently localize and recognize arbitrary regions of interest. It demands a single vision model that is capable of understanding the regional context and simultaneously executing perception tasks such as segmentation, recognition, and captioning. 
However, existing models often focus on either localizing class-agnostic masks, \eg, SAM~\cite{sam} and its efficiency-oriented followups \cite{zhang2023faster,xiong2023efficientsam,sun2024vrp}, or extracting only visual semantics, \eg, CLIP~\cite{radford2021learning} and its region-level variants \cite{zhong2022regionclip,yang2023recognize,sun2023alpha}.
Specifically, SAM develops a segmentation foundation model that can segment anything via prompting, enabling strong generalization in pixel-wise localization tasks. 
On the other hand, CLIP trains a recognition foundation model via contrastive learning on web-scale image-text pairs, demonstrating powerful zero-shot abilities in recognition tasks. 
Accordingly, learning semantic priors from a CLIP model within SAM's architecture offers a promising pathway towards comprehensive visual perception.

Our primary goal is to build a unified, promptable model that can segment, recognize, and caption anything at once (\cref{fig:insight}a). However, building such a foundation model is non-trivial, as 1) there is \textbf{no \textit{promptable} framework} capable of achieving generalist perceptions, 2) there is currently \textbf{no \textit{public} web-scale dataset} with paired masks, classes, and captions, and 3) there is \textbf{no training method} that can effectively and efficiently integrate the capabilities of CLIP and SAM into a single model for understanding arbitrary regions. This work carefully explores this direction and aims to provide a systematic solution that includes a novel framework, a new dataset, and an effective learning method. 

We start by introducing a promptable tokenization and captioning framework (\cref{fig:overview}), that enables simultaneous segmentation, recognition, and captioning. This requires a unified model capable of abstracting general-purpose representations, \eg, mask tokens and semantic tokens, given flexible prompting that cues any region of interest. 
We follow SAM's architecture, but upgrade its mask decoder into a generic image decoder, where an additional semantic token is produced for each predicted mask. The mask token contributes to  pixel-wise segmentation, similar to SAM, while the semantic token is responsible for region-level semantic prediction. By leveraging the semantic token, the model can concurrently address the open-vocabulary classification task through an MLP head, and the promptable captioning task with a lightweight text decoder using an auto-regressive process. We refer to this model as TAP, an acronym  for \textbf{T}okenize \textbf{A}nything via \textbf{P}rompting, as illustrated in \cref{fig:insight}b. 

Training such a highly performant and generalizable model necessitates a diverse, large-scale dataset. Nevertheless, there is currently no web-scale data source available for simultaneous segmentation and recognition. SA-1B~\cite{sam} constructs 1.1B high-quality mask annotations on 11M images for training a segmentation foundation model, \eg, SAM. Conversely, LAION-2B~\cite{schuhmann2022laion} collects 2B image-text pairs from the web, facilitating the training of generalizable recognition models, \eg, CLIP.
To address the challenge posed by the lack of aligned data, we introduce the \textit{SemanticSA-1B} dataset (see \cref{fig:insight}c). This dataset implicitly integrates web-scale semantics from LAION-2B into SA-1B. Specifically, for each segmented region within SA-1B, we extract its concept distribution over a concept vocabulary as its semantic prior, which is predicted by a powerful CLIP model trained on massive LAION image-text pairs. As a result, the SA-1B data, along with LAION-2B priors, contribute to our pre-training dataset.
\begin{figure*}[ht!]
\begin{center}
\includegraphics[width=0.49\linewidth]{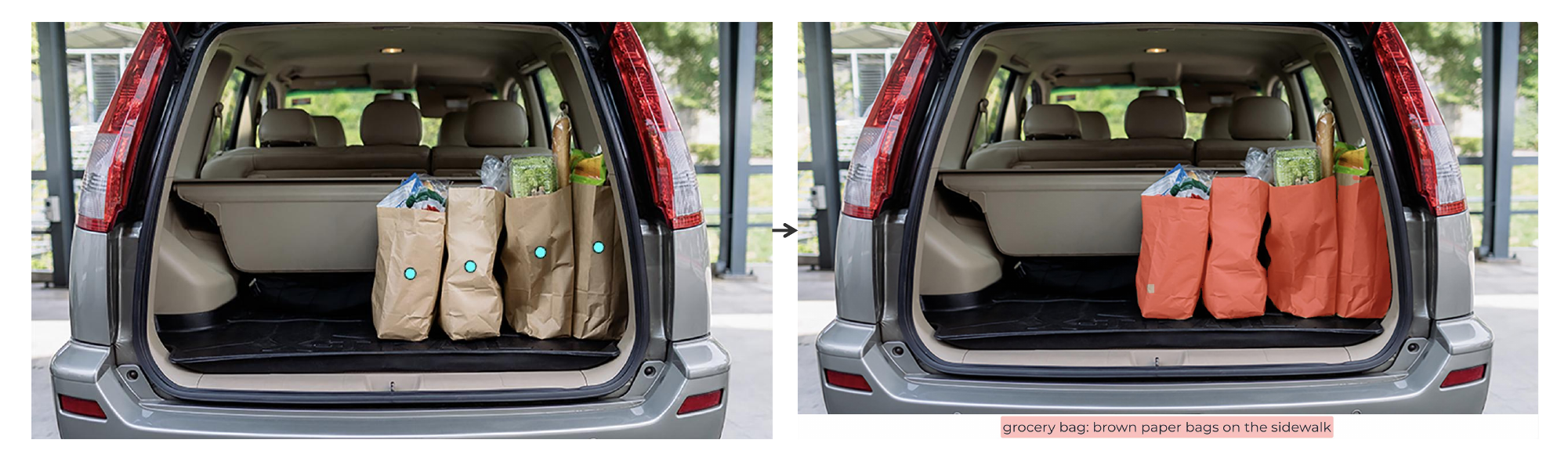}
\includegraphics[width=0.49\linewidth]{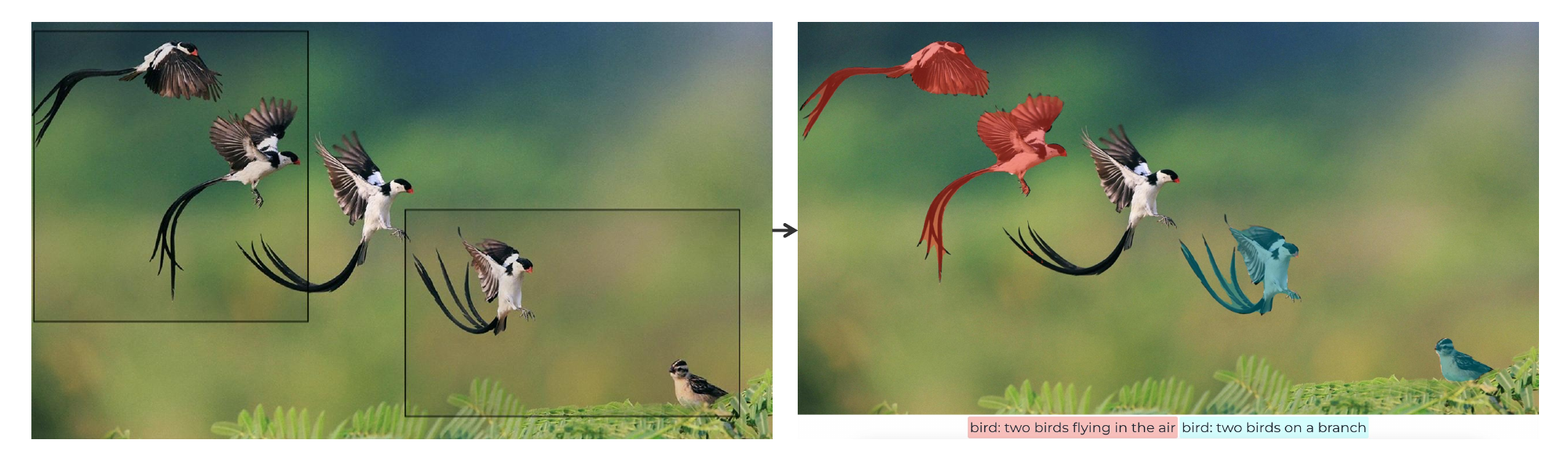}
\includegraphics[width=0.49\linewidth]{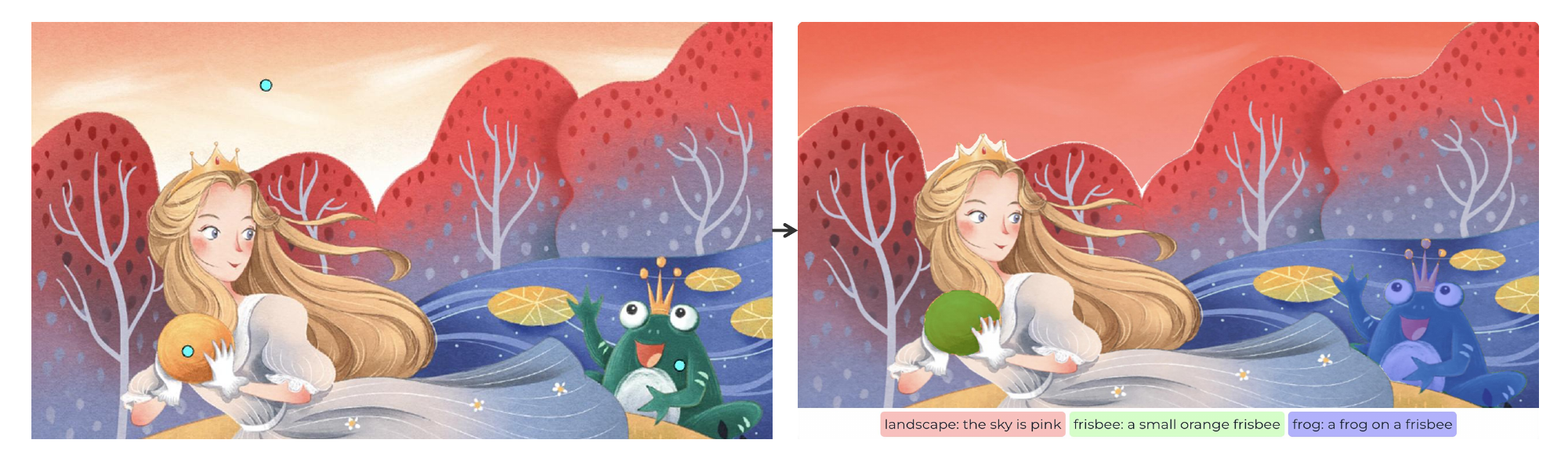}
\includegraphics[width=0.49\linewidth]{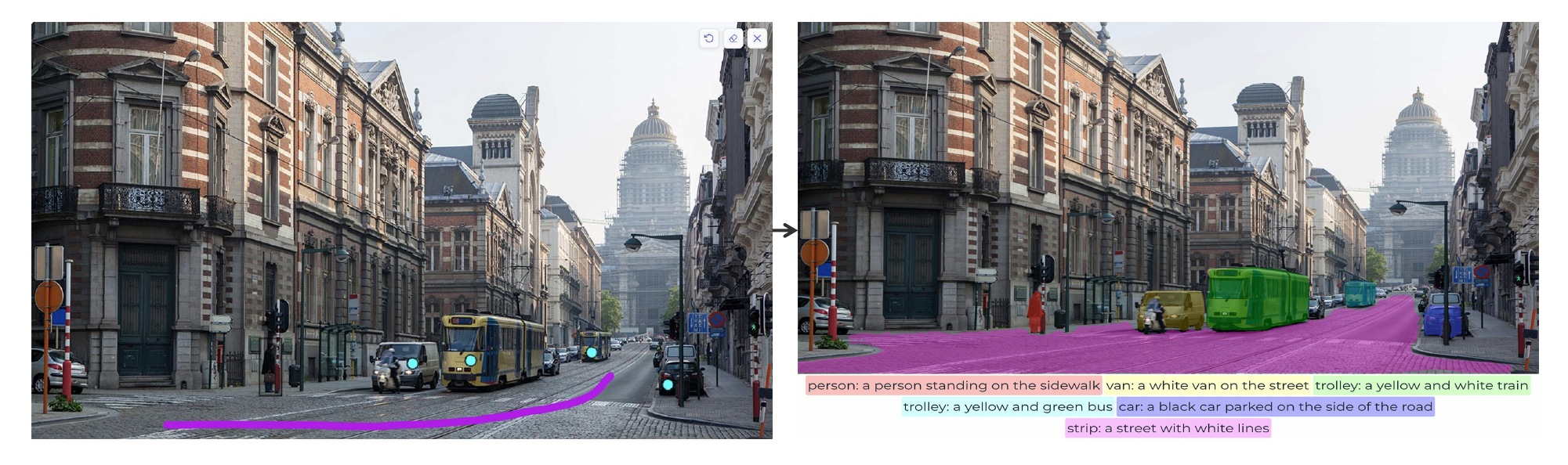}
\includegraphics[width=0.49\linewidth]{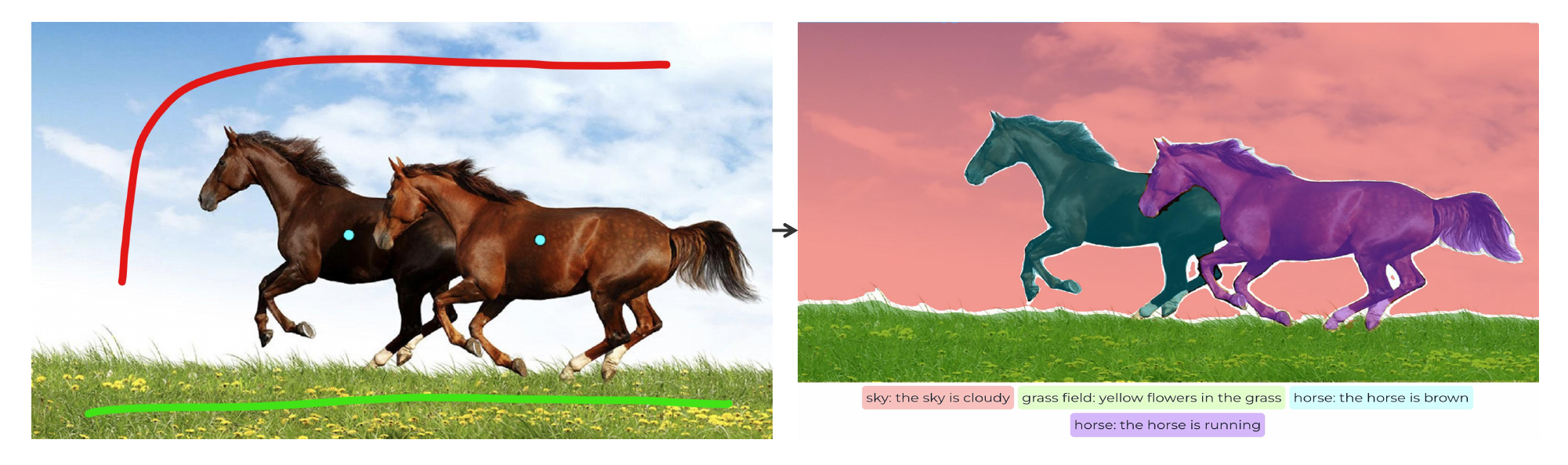}
\includegraphics[width=0.49\linewidth]{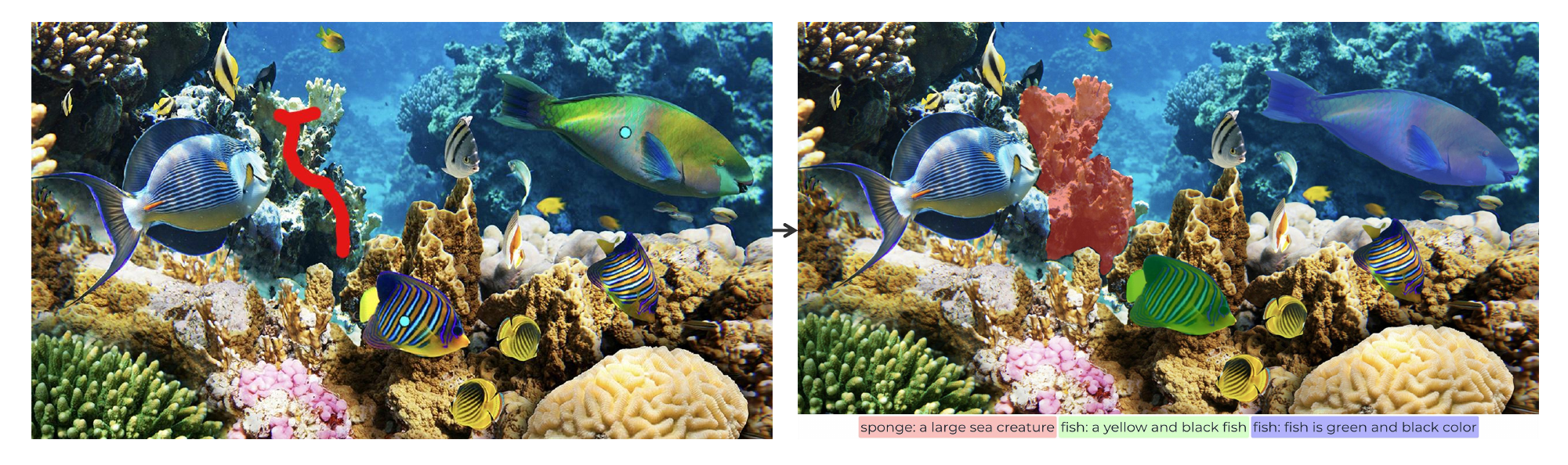}
\end{center}
\caption{\Ours accepts flexible prompts and outputs mask, category and caption at once.}
\label{fig:demo_1}
\end{figure*}

Using the \textit{SemanticSA-1B} dataset, we pre-train our model with ground-truth masks and associated semantics from the beginning, effectively integrating CLIP's capabilities within SAM's architecture. 
This is achieved simply by training a promptable tokenizer simultaneously for generic segmentation and concept prediction. To predict the semantic concept for each masked image, we further propose minimizing the KL divergence loss between the predicted concept distribution and the target distribution, aiming to maximize the transfer of CLIP knowledge.
This joint training objective enables powerful generalization in both localization and recognition, thereby facilitating general-purpose vision tasks.

We extensively evaluated our TAP model and its components. 
TAP demonstrates strong zero-shot performance in instance classification, \eg, 59.1 AP on the challenging LVIS~\cite{gupta2019lvis} benchmark, while maintaining competitive zero-shot segmentation performance, \eg, 43.0 vs.\ 43.1 AP for \Ours and SAM. 
Notably, we set a new record with a CIDEr score of 164.7 in the region caption task on Visual Genome~\cite{krishna2017visual}, using significantly fewer parameters compared to the prior works. 
Our findings indicate that the tokenized region-level features are generalizable for both segmentation and classification tasks, and can even directly prompt the causal text generation. Above all, we believe \Ours model can be a versatile region-level image tokenizer, capable of encoding regional context for a broad range of visual perception tasks (see \cref{fig:demo_1}).

%% file: arxiv_v2/2_related_works.tex
\section{Related work}\label{sec:formatting}

\subsection{Vision Foundation Models}

Vision foundation models aim to achieve strong zero and few-shot generalization capabilities across a broad range of vision tasks. 
Starting with CLIP \cite{radford2021learning}, which simultaneously trains image and text encoders with massive image-text pairs to align two
modalities, numerous efforts have emerged to train a general-purpose vision-language representation at scale~\cite{jia2021scaling,li2023scaling,sun2023evaclip}.
In addition, some works aim to build vision generalist models \cite{Painter, wang2023seggpt,sam,seem,asm}. For example, SAM \cite{sam} introduces a large-scale dataset and trains a model for promptable segmentation. Taking user interactions as prompts, SAM demonstrates strong zero-shot performance in general segmentation tasks. Concurrent to SAM, SegGPT \cite{wang2023seggpt} unifies a variety of segmentation tasks into one in-context segmentation problem. SegGPT showcases the capability to execute arbitrary segmentation tasks through in-context inference. Some other works seek to build a generalist model by leveraging multi-modality datasets \cite{alayrac2022flamingo,lu2022unified,sun2023generative}. In this work, we aim to build a vision foundation model that serves as a versatile region-level image tokenizer, capable of encoding general-purpose region context for a broad range of perception tasks.

\subsection{Open-Vocabulary Segmentation}
Unlike previous instance segmentation and semantic segmentation models~\cite{ he2017mask,bolya2019yolact,SOLO,SOLOV2, long2015fully,xiao2018unified,maskformer} that work in a limited vocabulary, open-vocabulary segmentation (OVS) aims to classify regions that go beyond the closed-vocabulary used for training \cite{LSeg, ODISE, FreeSeg, X-Decoder, MaskCLIP, MaskCLIP+, OVSeg, OpenSeg, zegformer, XPM, OpenSeeD,groupvit}. Numerous efforts focus on leveraging pre-trained Vision-Language models (VLMs) like CLIP \cite{radford2021learning} and center on designing specific alignment techniques to effectively integrate VLM knowledge into existing segmentation models~\cite{zegformer, LSeg, MaskCLIP, MaskCLIP+}. For example, LSeg~\cite{LSeg} embeds text and pixel embeddings into a common feature space, assigning label to each pixel. MaskCLIP \cite{MaskCLIP} builds a two-stage model to seamlessly integrate with CLIP visual encoder. ZegFormer \cite{zegformer} decouples problem into a class-agnostic grouping task and a region-level classification task to utilize VLM.
By leveraging the caption data, some studies align visual features with texts in a weakly supervised manner~\cite{OpenSeg,groupvit,OVSeg, CGG,XPM}. For instance, GroupViT~\cite{groupvit} is trained on image-caption pairs without pixel-level annotations, directly grouping masks based on text supervision. OVSeg~\cite{OVSeg} fine-tunes CLIP on masked images with pseudo labels generated from the nouns in image captions. CGG \cite{CGG}, on the other hand, combines grounding and generation losses to thoroughly explore the knowledge from image captions. 
Additionally, other studies~\cite{FreeSeg, X-Decoder,OpenSeeD} jointly learn multiple tasks within a single network or investigate text-to-image diffusion models \cite{ODISE,OVDiff}. 
Our work aligns with CLIP-based approaches but differs from two-stage models, which typically rely on an image-level CLIP to classify masks. Instead, our approach focuses on developing a single model with region-level semantic awareness.

\subsection{Zero-shot Region Understanding}

Previous works focus on extending the open-vocabulary capabilities of VLMs to object detection tasks~\cite{gu2021open,zareian2021open,zhong2022regionclip,kuo2022f,wang2023object}.
Recent studies~\cite{samclip,yang2023recognize} aim to merge CLIP's proficiency in open-vocabulary classification with SAM's capability in segmentation.  For instance, SAM-CLIP~\cite{samclip} distills knowledge from both SAM and CLIP by retraining the visual encoder with a portion of data sampled for two teachers, retaining the original strengths of both CLIP and SAM. RegionSpot~\cite{yang2023recognize} unifies prompting by adding an adapter trained on detection datasets, enabling SAM's mask tokens to interact with CLIP's features derived from masked image segments. Some works~\cite{semantic-sam, seem, asm} attempt to construct unified models capable of recognizing objects in arbitrary regions. SEEM~\cite{seem} was built upon X-Decoder~\cite{ X-Decoder}, excelling in handling various types of prompts, including clicks, bounding boxes, scribbles, text, and referring image segments. Following SAM~\cite{sam}, ASM~\cite{asm} created a new dataset (AS-1B) for SA-1B(~\cite{sam}), constructing rich annotations of semantic tags, question-answering pairs, and detailed captions. Leveraging this dataset, they develop a new model, ASM, for panoptic visual recognition. Unlike these models relying on handcrafted multi-modal datasets, we fully leverage extensive segmentation masks from SA-1B and semantic priors from a high-performing CLIP model, aiming to develop a promptable tokenizer that can understand semantic context for any given region.

%% file: arxiv_v2/3_method.tex
\section{Approach}
\vspace{-4pt}
\begin{figure}[tb]
\centering
\includegraphics[width=0.96\linewidth]{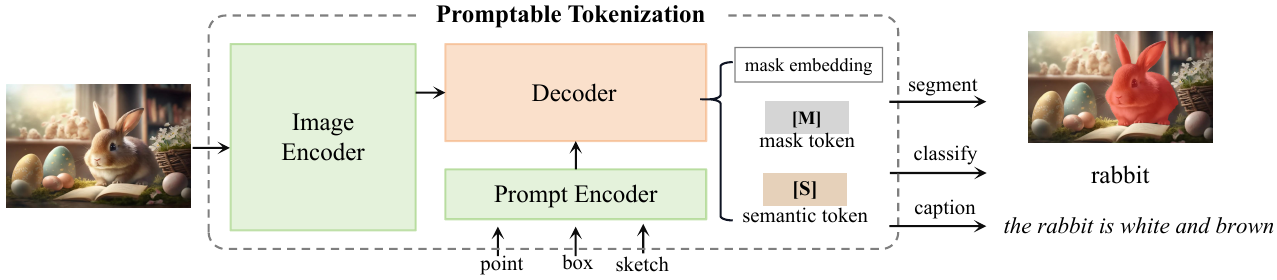}
\caption{\small Overview of \textbf{\Ours}. a) Building upon SAM's architecture, we enhance the mask decoder to a generic image decoder, adding an additional semantic token [S] to each predicted mask. b) Our model is pre-trained on SemanticSA-1B, jointly optimized for concept prediction and promptable segmentation. c) Subsequently, the pre-trained promptable tokenizer (in dotted box) is employed for region captioning.}
\label{fig:overview}
\end{figure}

\noindent We introduce a novel promptable framework that efficiently enables the segmentation, recognition, and captioning of arbitrary regions of interest. This is achieved by pre-training a promptable tokenizer that utilizes extensive segmentation masks with CLIP priors (Sec.~\ref{sec:token}), and subsequently expanding the model's capabilities to include generative abilities for promptable captioning (Sec. \ref{sec:caption}).

\subsection{Promptable Tokenization}\label{sec:token}
Our primary objective is to align vision and language within a promptable segmentation model to enhance the model with region-level semantic awareness. To achieve this goal, we introduce our model architecture, pre-training dataset, and learning method involving concept prediction and promptable segmentation, as well as the pre-training loss, in this subsection.

\paragraph{\textbf{Model architecture.}}
Our tokenizer model comprises three essential modules: an image encoder, a prompt encoder, and an image decoder (see \cref{fig:overview}). We maintain SAM's architecture but upgrade its mask decoder to a generic image decoder. Additionally, to more efficiently and effectively achieve our objectives, we make several modifications to SAM's architecture.  
Specifically, the image encoder adopts a standard Vision Transformer (ViT) \cite{dosovitskiy2020image}, where a 16$\times$16 non-overlapping window is employed. To alleviate computational intensity, we substitute the global attention in the image encoder with convolutional cross-window blocks~\cite{li2022exploring} and replace the query-based relative position embedding \cite{li2022mvitv2} with index-based relative position bias.
Regarding the prompt encoder, we do not add the mask prediction from the previous stage to the image embeddings, as it introduces discrepancies between the prior prompts (\eg., sketch points) and advanced prompts (\eg., interactive points). Consequently, all mask embedding layers in the prompt encoder are removed.
In the image decoder, we add an additional semantic token to each predicted mask, where the mask token is employed for pixel-wise segmentation, while the semantic token contributes to region-level recognition. Therefore, our image decoder produces a total of  4 masks and 9 tokens: 4 mask tokens, 4 semantic tokens, and an IoU token.

\paragraph{\textbf{Pre-training dataset.}}
Conventional vision-language alignment methods rely on image-text pairs \cite{sharma2018conceptual,changpinyo2021cc12m,schuhmann2022laion}, limiting the fine-grained region understanding. In contrast to prior methods~\cite{seem,asm,minderer2023scaling} reliant on well-collected or approximated region-text data, we align image segments with language using only segmentation data and CLIP priors. 
As SA-1B is a class-agnostic dataset, we utilize the high-performing open-source CLIP model, EVA-CLIP \cite{sun2023evaclip}, to compute the concept distribution $P_{\text{target}}$ as the semantic prior for each image segment within SA-1B. 
We first create a label list consisting of 2560 categories collected from various popular image datasets. Then, we employ a simple prompt template: `a $\left\{\right\}$' or `a photo of a $\left\{\right\}$' to generate the text embeddings $T_{C}$ using CLIP. Meanwhile, for each masked image segment from SA-1B, we obtain its visual embeddings $V_{C}$ generated by CLIP.  The concept distribution can be defined as follows:
\begin{equation} \label{eq:1}
P_{\text{target}}=\text{Softmax}(\frac{V_{C} \cdot T_{C} /\tau}{\left \| V_{C}  \right \| \cdot \left \| T_{C} \right \| })
\end{equation}

\noindent Here, $\tau$ denotes temperature parameter. Consequently, the segmentation data along with its off-the-shelf CLIP priors ($V_{C}, T_{C}, P_{\text{target}}$,) are stored locally, constituting  our pre-training dataset, SemanticSA-1B.

\paragraph{\textbf{Concept prediction.}} 
To enhance  our model with semantic awareness, we propose to predict region concepts using the semantic token. Concretely, we employ the semantic token to obtain the predicted visual embedding $V_{P}$, which is further projected to the concept distribution $P_{\text{pred}}$,
\begin{equation} \label{eq:2}
P_{\text{pred}}=\text{Softmax}(\frac{V_{P} \cdot T_{C} /\tau}{\left \| V_{P}  \right \| \cdot \left \| T_{C} \right \| })
\end{equation}

\noindent We further propose to align the concept distribution between model's prediction and CLIP's target. The concept alignment loss can  be defined as the KL divergence loss between $P_{\text{pred}}$ and $P_{\text{target}}$, represented as
\begin{equation} \label{eq:3}
\mathcal{L}_{\text{concept}}=\mathcal{L}_{\text{KL}}(P_{\text{pred}}\left|\right| P_{\text{target}})
\end{equation}

\noindent Different from feature alignment that typically minimizes the negative cosine similarity between the predicted visual embedding and CLIP visual embedding, formed as $\mathcal{L}_{\text{feat}}$$=$$-\frac{V_{P} \cdot V_{C}}{\left \|V_{P} \right \|_{2}\cdot \left \|V_{C} \right \|_{2}}$, concept alignment minimizes $\mathcal{L}_{\text{concept}}$ between two distributions. It measures the similarity between $V_{P}$ and $T_{C}$, drawing $V_{P}$ closer to positive $T_{C}$ (\ie, relevant concepts), while pushing it away from negative $T_{C}$ (\ie, irrelevant concepts). This encourages $V_{P}$ to be \textit{orthogonal}, maximizing the transfer of CLIP's open-world knowledge. 

\paragraph{\textbf{Promptable segmentation.}} 
The mask decoder within SAM responses to input prompts for generic segmentation. We thus consider promptable segmentation as a necessary prelude to unsealing the semantic capabilities. Following SAM, our model defaults to predicting four masks for each prompt, yet a routing strategy selects one to resolve the ambiguity.
To improve training efficiency on the large-scale SA-1B dataset, we implement a two-stage sampling strategy with maximal 9 prompt points, as it is performed within 11 interactive stages in the original SAM. In the first stage, we sample a box or point with equal probability from the ground-truth mask. In the subsequent stage, we uniformly sample 1 to 8 points from the error region between predicted and ground-truth masks.
To enable mask as the prior prompt, an aspect unexplored in SAM, we introduce a non-interactive sampling method with a 50$\%$ probability in the second stage. This sampling uniformly fetches 1 to 9 points from the ground-truth mask, providing a wider prompt space.
Regarding segmentation loss $\mathcal{L}_{\text{seg}}$, we employ a linear combination of focal loss \cite{lin2017focal} , dice loss \cite{milletari2016fully}, and IoU prediction loss, weighted at a 20:1:1 ratio. The IoU prediction head is trained using  mean-square-error loss, supervised by the actual IoU  between the predicted mask and the ground truth mask, following SAM~\cite{sam}. 

\paragraph{\textbf{Pre-training loss.}} Our final pre-training loss is the joint loss combining both concept prediction and promptable segmentation: $\mathcal{L}$$ = $$\alpha\mathcal{L}_{\text{concept}}$+ $\beta\mathcal{L}_{\text{seg}}$, where the balance weights $\alpha,\beta$ are searched and empirically set to (1,1) to sufficiently learn rich CLIP semantics. Using  this joint loss, we train a promptable tokenizer on SemanticSA-1B. An overview of our method is illustrated in \cref{fig:overview}. 

\subsection{Promptable Captioning} \label{sec:caption}
\begin{figure}[tb]
\centering
\includegraphics[width=0.85\linewidth]{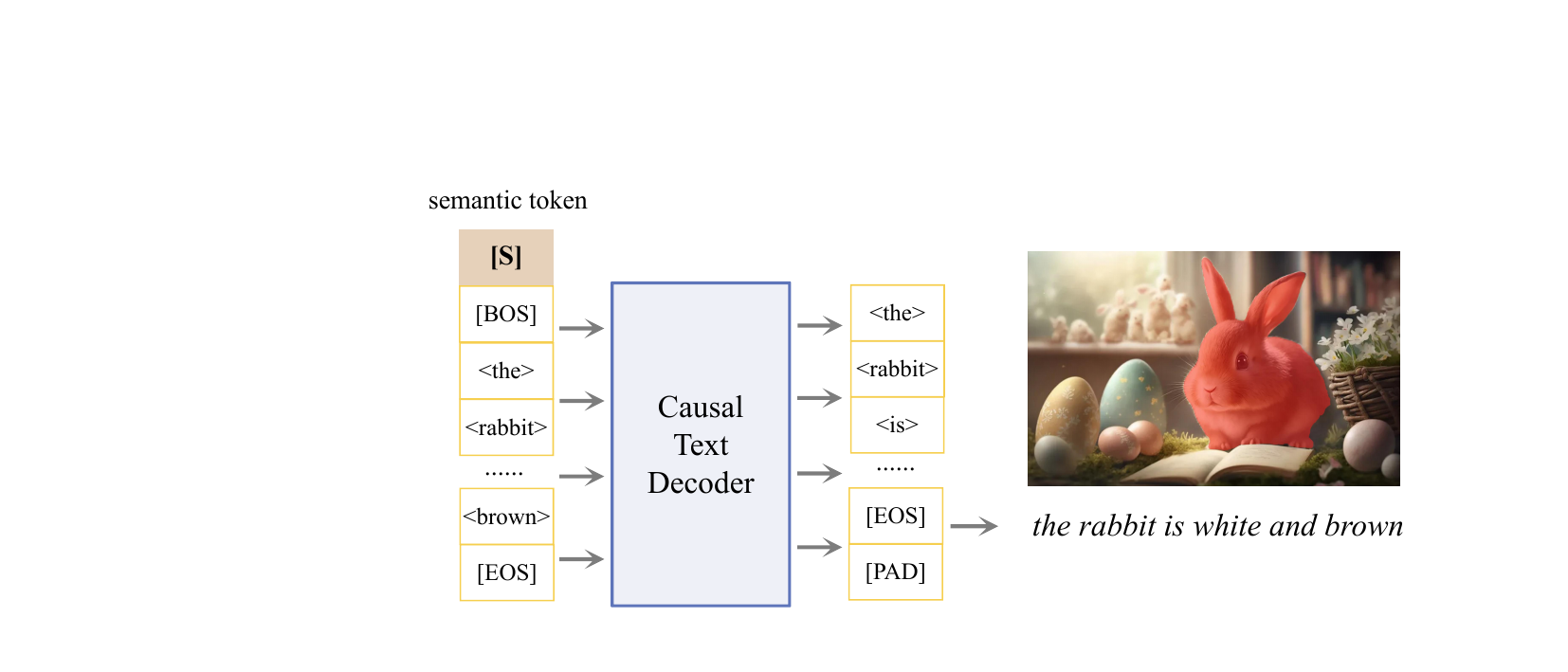}
\caption{\small\textbf{Promptable captioning.} Semantic token is used to prompt text generation.}
\label{fig:caption}
\end{figure}

In order to evaluate the effectiveness of the promptable semantic tokens, after pre-training on SemanticSA-1B, we append an additional lightweight text decoder at the top of the model, and fine-tune it on the Visual Genome (VG)~\cite{krishna2017visual} dataset. An overview of our text generation architecture is depicted in \cref{fig:caption}.
\paragraph{\textbf{Region caption task.}}
Many prior works \cite{wang2023caption,zhang2023gpt4roi,asm} generate region captions using CLIP visual features along with Large Language Models (LLMs). 
Recent methods \cite{wang2023caption,huang2023sca} utilize SAM decoder features also rely on LLMs to enhance the weak-semantic context.
However, simply appending LLMs not only imposes computational burdens but may also be unnecessary for the region-level visual understanding, as regional captions typically consist of fewer than 15 words.
In our effort to build the compact vision models, we develop a generative tokenizer. This is achieved by extending the capabilities of our tokenizer (Sec.~\ref{sec:token}) to incorporate text generation via causal language modeling.
Specifically, we train a lightweight text decoder prompted with semantic tokens from our tokenizer to generate the region caption. 
By leveraging this semantic-aware visual tokenizer, our model efficiently trains this task end-to-end, obviating the need for LLMs.

\paragraph{\textbf{Causal text decoder.}} 
We utilize a standard Transformer with an embedding dimension of 512 to yield the brief region descriptions. This lightweight text decoder is sufficient to perform mask-to-text translation if prompted with semantic context. Given semantic tokens generated by the promptable tokenizer (refer to \cref{fig:overview}), we solely apply a linear projection to these semantic tokens, aligning their dimensions with text embeddings (see \cref{fig:caption}).  
 Subsequently, we place the semantic token at the leading position of a sequence, followed by a \texttt{[BOS]} token and word tokens. Rotary embedding \cite{su2021roformer} is utilized to integrate the positional encoding for multi-modal sequences. We adopt byte-pair encoding \cite{sennrich2015neural} with a 32k token vocabulary. Eventually, we perform the next token prediction through causal language modeling, employing cross-entropy loss. 
 
\subsection{Inference}
After training visual perception on SemanticSA-1B and text generation on Visual Genome, our model is capable of conducting classification, segmentation, and captioning simultaneously. The following outlines the inference pipeline.

\paragraph{\textbf{Mask selection.}} Given a visual prompt, our image decoder produces 4 masks and 9 tokens. The final mask and the associated semantic token are selected using a heuristic strategy.
Specifically, we choose the first mask if prompted with a boundary box, and select the top-ranked remainder if prompted with loose points, akin to a simplified implementation of the mixture-of-experts (MoE) \cite{jacobs1991adaptive}.

\paragraph{\textbf{Concept prediction.}} The selected semantic token is then utilized for predicting concepts on a dataset-specific concept vocabulary (\eg, COCO and LVIS). Concretely, we employ the semantic token to obtain a 1024-dimension visual embedding through a 3-layer MLP (256$\rightarrow$1024$\rightarrow$1024). This visual embedding is further projected to the concept distribution logits (i.e., $P_{\text{pred}}$) for classification.

\paragraph{\textbf{Caption generation.}} We finally generate up to 40 word tokens with the greedy sampling strategy prompted by the selected semantic token. To speed up attention computation, we follow a standard practice for auto-regressive decoding,  caching the key and value pairs for the previous generation in the sequence.

%% file: arxiv_v2/4_experiments.tex
\section{Experiments} \label{sec:exps}
\subsection{Experiments Setup}

\paragraph{\textbf{Pre-training.}}
We pre-train \Ours models on SemanticSA-1B, which includes the SA-1B data along with their associated CLIP priors. The full SA-1B comprises 11 M high-resolution images with around 100 regions per image, totaling 1.1B segmentation masks. To obtain CLIP priors for the SA-1B data, inspired by \cite{yao2022detclip,zhong2022regionclip,minderer2023scaling}, we utilize EVA-CLIP \cite{sun2023evaclip} to generate text embeddings on a curated label space, merging from COCO \cite{coco}, ADE20K \cite{zhou2017scene}, LVIS \cite{gupta2019lvis}, Objects365 \cite{shao2019objects365}, Visual Genome \cite{krishna2017visual} and OpenImagesV4 \cite{kuznetsova2020open} datasets. This results in a concept list spanning 2560 categories, covering both \textit{things} and \textit{stuff} for segmentation.

\paragraph{\textbf{Evaluation.}}
We assess zero-shot instance segmentation performance on COCO and LVIS. For zero-shot instance classification, we prioritize LVIS due to its broader range of 1203 categories compared to COCO, which covers only 80 common categories, diverging from the open-world assumption. In the region-level captioning task, regarding the domain gap between SA-1B and Visual Genome (VG)~\cite{krishna2017visual}, we adopt a two-stage fine-tuning approach. We first freeze the image encoder-decoder and only fine-tune the text decoder using the VG~\cite{krishna2017visual} v1.0 train set, denoted as  `partial-FT'. Subsequently, we unfreeze the image encoder-decoder and fine-tune the model end-to-end.  We mark this two-stage fine-tuning strategy  as `full-FT'. Since there is no web-scale dataset with aligned masks, classes, and captions, all the ablation studies (Sec.~\ref{sec:abl}) are conducted with `partial-FT'. We report the following four metrics on the VG test set and RefCOCOg \cite{mao2016generation} validation set: BLEU@4, METEOR, ROUGE, and CIDEr. 

\paragraph{\textbf{Implementation details.}}
 We utilize the AdamW \cite{loshchilov2019decoupled} optimizer ($\beta_{1}$ = 0.9, $\beta_{2}$ = 0.999) with a base learning rate of 1e-3 in all experiments. A cosine learning rate schedule \cite{loshchilov2017sgdr} is implemented. During pre-training on SemanticSA-1B, scale jitter \cite{ghiasi2021simple} is applied with a range of [0.5, 2.0] for 180k iterations ($\sim$4 epochs), with a batch size of 256 across 256 GPUs. We fine-tune VG without data augmentation for 60k iterations ($\sim$50 epochs), with a batch size of 64 across 64 GPUs. Additional hyper-parameters include a weight decay of 0.1, a drop path \cite{huang2016deep} rate of 0.1/0.2 for ViT-B/ViT-L, and a dropout \cite{srivastava2014dropout} rate of 0.1/0.4 for the image/text decoder. The image encoder is initialized from MAE \cite{he2022masked} pre-trained weights, while all other layers are from scratch.
For all experiments, we adopt up to 64 sampled prompts per GPU at each sampling stage.

\subsection{Main Results}

\paragraph{\textbf{Zero-shot instance classification.}}
We prompt our model with ground-truth (GT) boxes to evaluate the \textit{bare} recognition capability on LVIS. With GT boxes as visual prompts, our model substantially surpasses RegionCLIP~\cite{zhong2022regionclip} and RegionSpot~\cite{yang2023recognize}, which are trained on limited image regions. These promising results suggest that employing concept prediction on exhaustive image regions can effectively empower SAM with semantic awareness. As shown in~\cref{tab:zsic}, the highly performant EVA-CLIP outperforms all other methods in zero-shot evaluation, achieving an impressive rare AP. Nonetheless, deploying a standalone CLIP (5B) model to compute massive image crops is impractical for real-time vision systems. We demonstrate that the  knowledge of large CLIP models can be integrated into a compact tokenizer (0.1B) with acceptable performance.

\begin{table}[ht]
\caption{\small
\textbf{Zero-shot instance classification} on LVIS~\cite{gupta2019lvis}. 
All entries are evaluated using GT boxes for a fair comparison. Superscripts `R', `C', `F' refer to rare, common and frequent categories as defined in LVIS evaluation.}
\label{tab:zsic}
\centering
\setlength{\tabcolsep}{3mm}
\renewcommand\arraystretch{1}
\resizebox{0.96\linewidth}{!}{
\begin{tabular}{l|c|c|cccc}
\toprule
Model & Params & Training data & AP & AP$^{R}$ & AP$^{C}$ & AP$^{F}$ \\
\midrule
\multicolumn{5}{l}{\textit{Supervised detector:}} \\
ViTDet-B \cite{li2022exploring} & 0.1B & LVIS & \textcolor{gray!70}{61.9} & \textcolor{gray!70}{40.8} & \textcolor{gray!70}{58.5} & \textcolor{gray!70}{74.9} \\
ViTDet-L \cite{li2022exploring} & 0.3B & LVIS & \textcolor{gray!70}{68.8} & \textcolor{gray!70}{51.5} & \textcolor{gray!70}{65.6} & \textcolor{gray!70}{79.9} \\
\midrule
\multicolumn{5}{l}{\textit{Image-level CLIP:}} \\
CLIP-L \cite{radford2021learning} & 0.3B & WIT-400M & 48.8 & 52.8 & 50.0 & 45.6 \\
EVA-CLIP-E \cite{sun2023evaclip}  & 5B   & LAION-2B & 64.3 & 72.4 & 65.3 & 59.7 \\
\midrule
\multicolumn{5}{l}{\textit{Region-level CLIP:}} \\
RegionCLIP-R50x4 \cite{zhong2022regionclip} & 0.1B & CC-3M       & 50.7 & 50.1 & 50.1 & 51.7 \\
RegionSpot-BL \cite{yang2023recognize}      & 0.4B & O365,OI,V3D & 56.6 & 50.6 & 50.2 & 68.8 \\
\midrule
\multicolumn{5}{l}{\textit{Promptable tokenizer:}} \\
\Ours-B                    &  0.1B & SemanticSA-1B & 57.4 & 58.6 & 56.8 & 57.5 \\
\rowcolor{pink!15} \Ours-L & 0.3B  & SemanticSA-1B & 59.1 & 61.7 & 58.9 & 58.3 \\
\bottomrule
\end{tabular}}
\end{table}

\paragraph{\textbf{Region-level captioning.}}
We assess our model on Visual Genome \cite{krishna2017visual} and RefCOCOg \cite{mao2016generation}. Initially, we utilize GT boxes to prompt the image decoder, and subsequently, we employ the resulting semantic tokens to prompt the text decoder. The evaluation results are presented in \cref{tab:regioncap}. Surprisingly, our model achieves a CIDEr score of 154.7 on Visual Genome, even with a frozen image encoder-decoder that is pre-trained on SA-1B and has not seen VG images before (`partial-FT'). By adopting a two-stage fine-tuning strategy (`full-FT'), we set a new record with a CIDEr score of 164.7, only using a lightweight text decoder. It is noteworthy that the concurrent work ASM~\cite{asm} is trained on a multi-modal dataset, including a vast repository of region-text pairs. Semantic knowledge of our model is learnt from a CLIP model. Another concurrent work, SCA \cite{huang2023sca}, additionally trains a 12-layer image decoder to learn caption tokens for text prompting. These results suggest that our semantic token effectively encodes sufficient region-level information during pre-training for captioning, supporting  our earlier claim that \Ours can function as a location-aware image tokenizer.
\begin{table}[ht]
\caption{\small
\textbf{Region captioning} on Visual Genome~\cite{krishna2017visual} and RefCOCOg~\cite{mao2016generation}. GT boxes are utilized as as the region proposals.}
\label{tab:regioncap}
\centering
\setlength{\tabcolsep}{0.7mm}
\renewcommand\arraystretch{1.3} 
\resizebox{0.96\linewidth}{!}{
\begin{tabular}{lcllcccc}
\toprule
\multirow{2}[0]{*}{} & 
\multicolumn{3}{c}{} & \multicolumn{2}{c}{\textbf{Visual Genome}} & \multicolumn{2}{c}{\textbf{RefCOCOg}} \\
\cmidrule(lr){5-6}\cmidrule(lr){7-8}
Method & VisualEncoder & TextDecoder & TextPrompt & METEOR & CIDEr & METEOR & CIDEr \\
\midrule 
GRiT\cite{wu2022grit}          & ViT-B  & Small-43M                       & BoxFeature & 17.1 & 142.0 & 15.2 & 71.6 \\
GPT4ROI\cite{zhang2023gpt4roi} & CLIP-H & Vicuna7B\cite{zheng2023judging} & BoxFeature & 17.4 & 145.2 & - & - \\
ASM\cite{asm}                  & ViT-g  & Husky7B\cite{liu2023internchat} & BoxFeature & 18.0 & 145.1 & 20.8 & 103.0 \\
AlphaCLIP\cite{sun2023alpha}   & ViT-L  & Vicuna7B\cite{zheng2023judging} & BoxFeature &  18.9 & 160.3 & 16.7 & 109.2 \\
SCA\cite{huang2023sca}         & SAM-H  & Llama3B\cite{geng2023openllama} & \small{CaptionToken} & 17.4 & 149.8 & 15.6 & 74.0 \\
\midrule
\Ours (partial-FT)                 & ViT-B & Small-38M & SemanticToken & 17.7 & 152.3 & 19.2 & 93.5 \\
\Ours (partial-FT)                 & ViT-L & Small-38M & SemanticToken & 17.9 & 154.7 & 19.5 & 95.6 \\
\rowcolor{pink!15} \Ours (full-FT) & ViT-L & Small-38M & SemanticToken & 18.9 & \textbf{164.7} & 20.4 & 102.8 \\
\bottomrule
\end{tabular}}
\end{table}

\paragraph{\textbf{Zero-shot instance segmentation.}}
We evaluate our model in zero-shot instance segmentation, a task at which the original SAM excels.  Following a common practice~\cite{sam, yang2023recognize}, we first obtain detection bounding boxes from a ViTDet-H model~\cite{li2022exploring}. Subsequently, we utilize these boxes to prompt the image decoder and compare the \textit{bare} segmentation performance (\ie, using the box category) on COCO and LVIS. For a fair comparison, we report results from both the original SAM and our reproduced version (denoted as \textit{our impl}.). As depicted in \cref{tab:zsis}, our model achieves comparable segmentation results with original SAM across different model scales. This demonstrates that additional concept prediction and region captioning tasks do not compromise SAM's original capability. Moreover, it suggests that generic segmentation, being an elementary and geometric task, may not fully exploit the semantic representation in vision foundation models.

\begin{table}[ht] 
\caption{\small
\textbf{Zero-shot instance segmentation} on COCO~\cite{coco} and LVIS~\cite{gupta2019lvis}. 
Box proposals are obtained using ViTDet-H\cite{li2022exploring}, a typical supervised detection method.
}
\label{tab:zsis}
\centering
\setlength{\tabcolsep}{2mm}
\renewcommand\arraystretch{1.3} 
\resizebox{0.96\linewidth}{!}{ 
\begin{tabular}{lcccc|ccccccc}
\toprule
\multirow{2}[0]{*}{} & \multicolumn{4}{c}{\textbf{COCO}} & \multicolumn{7}{c}{\textbf{LVIS}} \\
Model & AP & AP$^{S}$ & AP$^{M}$ & AP$^{L}$ & AP & AP$^{S}$ & AP$^{M}$ & AP$^{L}$ & AP$^{R}$ & AP$^{C}$ & AP$^{F}$ \\
\midrule
ViTDet-H \cite{li2022exploring} & \textcolor{gray!70}{51.0} & \textcolor{gray!70}{32.0} & \textcolor{gray!70}{54.3} & \textcolor{gray!70}{68.9} & \textcolor{gray!70}{46.6} & \textcolor{gray!70}{35.0} & \textcolor{gray!70}{58.0} & \textcolor{gray!70}{66.3} & \textcolor{gray!70}{35.9} & \textcolor{gray!70}{46.8} & \textcolor{gray!70}{51.1} \\
\midrule
SAM-B \cite{sam} & 41.1 & 28.3 & 45.6 & 53.7 & 40.8 & 30.1 & 53.0 & 58.5 & 32.6 & 41.9 & 43.3 \\
SAM-L \cite{sam} & 45.5 & 30.2 & 50.1 & 60.4 & 43.8 & 31.9 & 56.7 & 64.2 & 34.3 & 44.7 & 46.9 \\
SAM-H \cite{sam} & 46.5 & 30.8 & 51.0 & 61.7 & 44.7 & 32.5 & 57.6 & 65.5 & 34.6 & 45.5 & 47.8 \\
\midrule
SAM-B (\footnotesize our impl.)  & 45.1 & 28.1 & 50.1 & 61.4 & 42.1 & 29.3 & 54.9 & 64.2 & 33.2 & 43.2 & 44.7 \\
SAM-L (\footnotesize our impl.)  & 46.0 & 29.0 & 50.7 & 62.2 & 43.1 & 30.2 & 56.0 & 65.3 & 33.4 & 44.2 & 46.1 \\
\midrule
\Ours-B                    & 45.1 & 28.7 & 50.1 & 60.6 & 42.2 & 29.5 & 55.0 & 64.0 & 33.6 & 43.1 & 45.0 \\
\rowcolor{pink!15} \Ours-L & 46.0 & 29.1 & 50.9 & 62.2 & 43.0 & 30.2 & 55.9 & 65.1 & 33.7 & 44.1 & 46.0 \\
\bottomrule
\end{tabular}}
\end{table}

\subsection{Ablation Study} \label{sec:abl}
\paragraph{\textbf{Pre-training loss.}}
Ablation studies on pre-training loss are presented in \cref{tab:featdist,tab:maskcap}, where $\mathcal{L}_{\text{seg}}$, $\mathcal{L}_{\text{feat}}$ and $\mathcal{L}_{\text{concept}}$ represent pre-training with segmentation, feature prediction and concept prediction respectively (Sec.~\ref{sec:token}).
As observed in \cref{tab:maskcap}, caption metrics are remarkably low when pre-trained with  $\mathcal{L}_{\text{seg}}$ alone (Model A). When combined with semantic prediction (Model B/C), caption performance sees a significant improvement. Despite showing semantic awareness, feature prediction is inferior to concept prediction in both  classification and captioning tasks. These findings indicate that a concept space is crucial for acquiring CLIP priors. We conjecture that this space has efficiently facilitated the model's learning of negative text embeddings (i.e., $T_{C}$) from CLIP. In addition, the segmentation results presented  in \cref{tab:maskcap} indicate that pre-training with additional semantic prediction neither enhances nor compromises the mask AP on COCO and LVIS. This observation also suggests that the SAM-wise architecture could incorporate more task supervision beyond the segmentation mask.

\begin{table*}[tb]
\caption{\small
\textbf{Ablation study on pre-training loss and text prompt.}
Default settings are marked in \colorbox{gray!15}{gray}.
}
\label{tab:maskcap} 
\centering
\setlength{\tabcolsep}{1mm}
\renewcommand\arraystretch{1.3} 
\resizebox{0.96\linewidth}{!}{
\begin{tabular}{lllcccccc}
\toprule
\multirow{2}[0]{*}{} &
\multicolumn{2}{c}{} & \multicolumn{4}{c}{\textbf{VG Caption}} & \multicolumn{2}{c}{\textbf{Segmentation}} \\
\cmidrule(lr){4-7}\cmidrule(lr){8-9} 
Model & Pre-train & TextPrompt & BLEU@4 & METEOR & ROUGE & CIDEr & AP$_\text{COCO}$ & AP$_\text{LVIS}$ \\
\midrule 
Model A                    & $\mathcal{L}_{\text{seg}}$         & MaskToken     & 8.8  & 13.2 & 29.0 & 105.2 & \textbf{46.1} & \textbf{43.2} \\
Model B                    & $\mathcal{L}_{\text{seg}}$,$\mathcal{L}_{\text{concept}}$  & MaskToken     & 11.1 & 16.6 & 34.0 & 138.9 & 46.0 & 43.0 \\
Model C                    & $\mathcal{L}_{\text{seg}}$,$\mathcal{L}_{\text{feat}}$ & SemanticToken & 11.4 & 16.9 & 34.7 & 143.1 & 46.0 & 43.1 \\
\rowcolor{gray!15} Model D & $\mathcal{L}_{\text{seg}}$,$\mathcal{L}_{\text{concept}}$  & SemanticToken & \textbf{12.4} & \textbf{17.9} & \textbf{36.2} & \textbf{154.7} & 46.0 & 43.0 \\
\bottomrule
\end{tabular}}
\end{table*}

\begin{table*}[tb]
\caption{
\textbf{Ablation study on semantic prediction tasks} for zero-shot classification. 
Default tasks are marked in \colorbox{gray!15}{gray}.
}\label{tab:featdist}
\centering
\setlength{\tabcolsep}{1mm}
\renewcommand\arraystretch{1.3} 
\resizebox{0.96\linewidth}{!}{
\begin{tabular}{clcccc|cccc}
\toprule
\multirow{2}[0]{*}{} & \multicolumn{1}{c}{} & \multicolumn{4}{c}{\textbf{COCO}} & \multicolumn{4}{c}{\textbf{LVIS}} \\
\cmidrule(lr){3-6}\cmidrule(lr){7-10} 
VisualEncoder & Pre-train & AP & AP$^{S}$ & AP$^{M}$ & AP$^{L}$ & AP & AP$^{R}$ & AP$^{C}$ & AP$^{F}$ \\
\midrule
ViT-L                     & $\mathcal{L}_{\text{seg}}$,$\mathcal{L}_{\text{feat}}$ & 62.0 & 44.5 & 69.9 & 75.4 & 39.1 & 35.5 & 37.4 & 42.7 \\
\rowcolor{gray!15} ViT-L  & $\mathcal{L}_{\text{seg}}$,$\mathcal{L}_{\text{concept}}$  & \textbf{77.0} (\textcolor{ForestGreen}{+15.0}) & \textbf{60.0} & \textbf{83.7} & \textbf{90.0} & \textbf{59.1} (\textcolor{ForestGreen}{+20.0}) & \textbf{61.7} & \textbf{58.9} & \textbf{58.3} \\
\bottomrule
\end{tabular}}
\end{table*}

\paragraph{\textbf{Semantic token.}}
To assess the effectiveness of semantic tokens, we conduct four experiments.  Firstly, we pre-train our model using the loss listed in the `Pre-train' column. Subsequently, we fine-tune the text decoder using the items outlined in `TextPrompt', generated from the frozen pre-trained model.  Model A serves as our baseline, pre-trained with only $\mathcal{L}_{\text{seg}}$. Here, mask tokens are directly used for region-level captioning task, akin to using the original SAM's output to train the text decoder. Model D is our default model, jointly optimized with promptable segmentation and concept prediction. Semantic tokens are used to prompt the text decoder. As demonstrated in \cref{tab:maskcap}, semantic tokens consistently outperform mask tokens in the captioning task while achieving comparable AP in the segmentation task. Eventually, the semantic token proves to be the most effective. This suggests that semantic tokenization significantly unlocks the potential of the foundation model, facilitating more perception tasks.

\begin{table*}[tb]
\caption{\small
\textbf{Ablation on text decoder architectures.}
The default is marked in \colorbox{gray!15}{gray}.
}\label{tab:scalingtext}
\centering
\setlength{\tabcolsep}{4mm}
\renewcommand\arraystretch{1.2}
\resizebox{0.96\linewidth}{!}{
\begin{tabular}{lccccccc}
\toprule
\multirow{2}[0]{*}{} & \multicolumn{3}{c}{\textbf{TextDecoder}} & \multicolumn{4}{c}{\textbf{VG Caption}} \\
\cmidrule(lr){2-4}\cmidrule(lr){5-8}
\footnotesize Model & \footnotesize Params & \footnotesize Depth & \footnotesize Dim & \footnotesize BLEU@4 & \footnotesize METEOR & \footnotesize ROUGE & \footnotesize CIDEr \\
\midrule 
\Ours-L                      & 20M & 6  & 512 & 12.0 & 17.6 & 35.9 & 153.2  \\
\Ours-L                      & 25M & 8  & 512 & 12.2 & 17.7 & 36.0 & 153.9  \\
\rowcolor{gray!15} \Ours-L   & 38M & 12 & 512 & \textbf{12.4} & \textbf{17.9} & \textbf{36.2} & \textbf{154.7} \\
\Ours-L                      & 43M & 6  & 768 & 12.3 & 17.8 & 36.0 & 154.2  \\
\bottomrule
\end{tabular}}
\end{table*}

\begin{figure*}[tb]
\begin{center}
\includegraphics[width=0.48\linewidth]{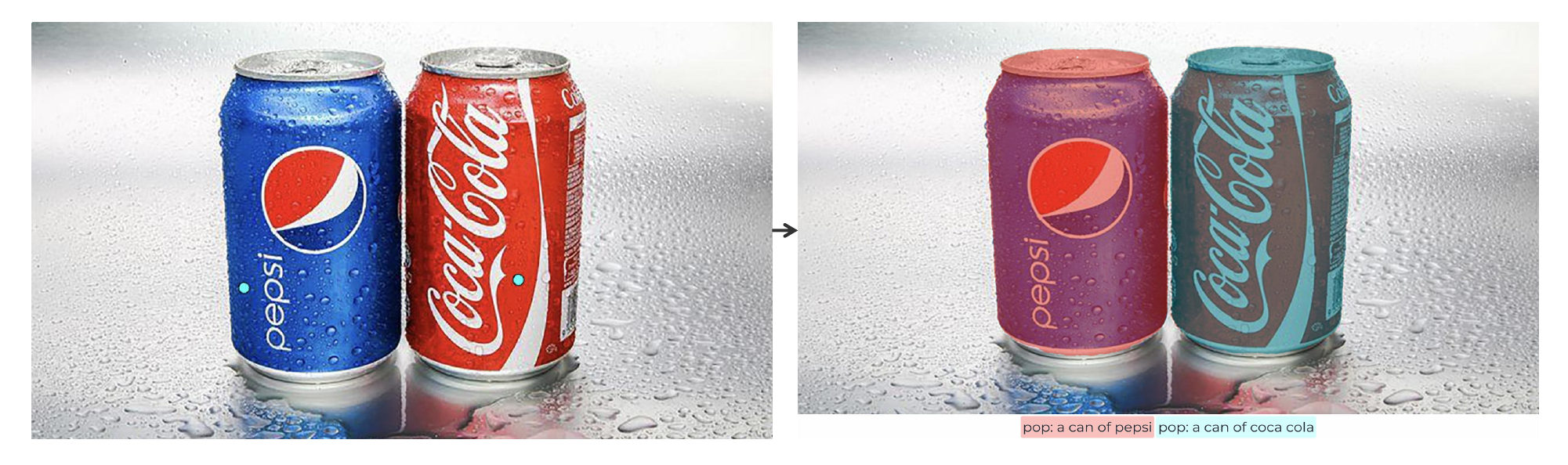}
\includegraphics[width=0.48\linewidth]{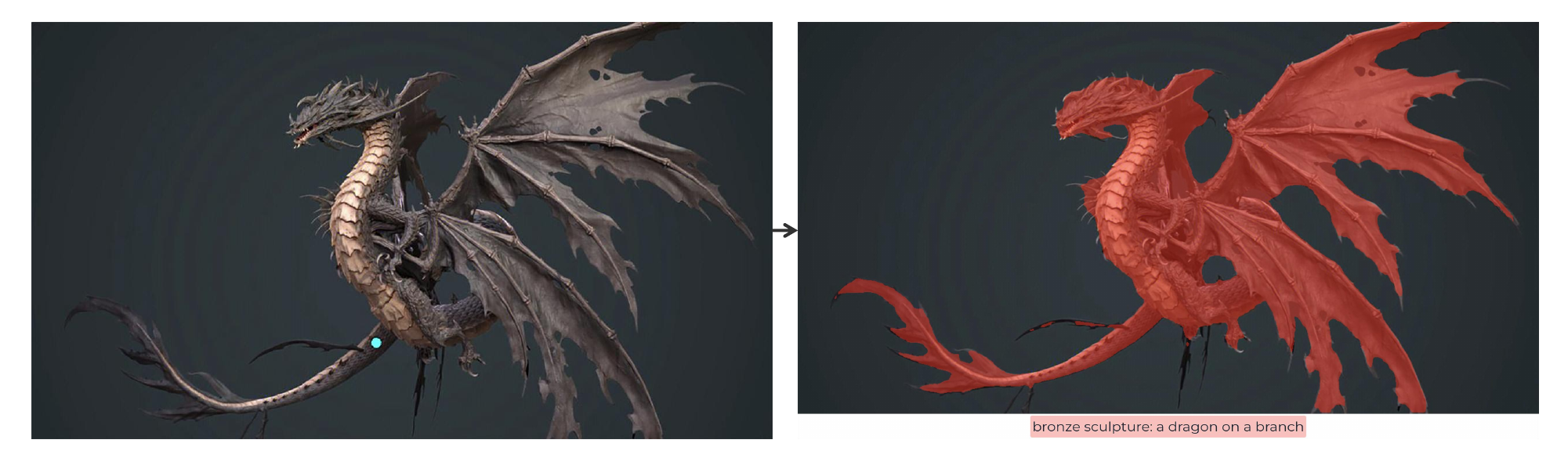}
\includegraphics[width=0.48\linewidth]{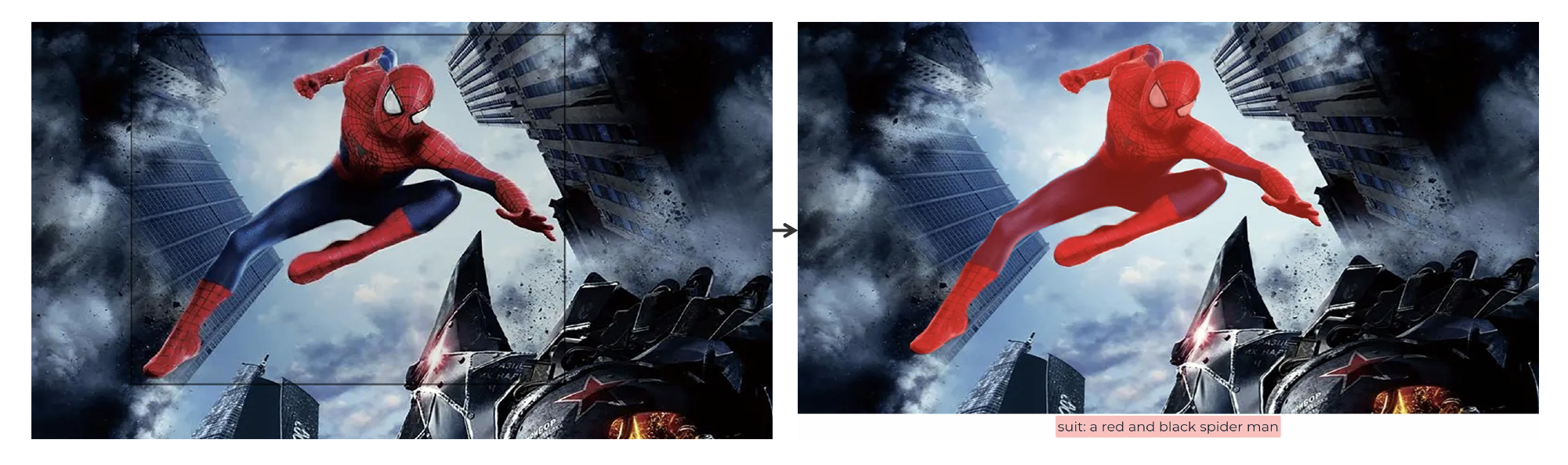}
\includegraphics[width=0.48\linewidth]{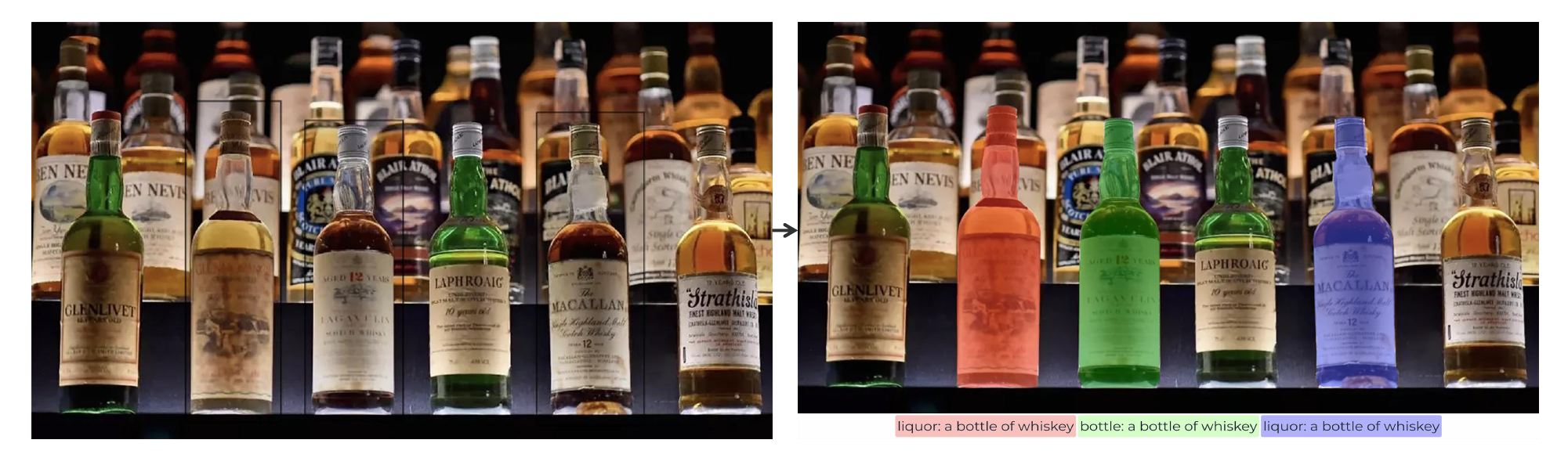}
\end{center}
\caption{\textbf{Visualization of understanding open-world knowledge.}}
\label{fig:demo_3}
\end{figure*}

\paragraph{\textbf{Scaling text decoder.}} We scale up the text decoder along the depth and embedding dimension to ablate the caption bottleneck. As illustrated in \cref{tab:scalingtext}, there is no substantial improvement with an increased model scale on the VG dataset. This suggests that employing larger decoders for region captioning may not be necessary unless the text length and quantity can be further increased.

\subsection{Qualitative Results}

We qualitatively evaluate \Ours using point-based prompts. By simply clicking or automatically prompting with a dense grid of points, our model can simultaneously generate the segmentation mask, category name, and text description.

\paragraph{\textbf{Open-world knowledge.}}
\cref{fig:demo_3} showcases example instances that pose challenges in open-world scenarios. Due to the subjective nature of vocabulary design, curated concepts such as ‘pepsi’, ‘cocacola’, ‘dragon’, ‘spider-man’ and ‘whisky’ could hardly be selected via retrievals (\ie, classification). However, our model demonstrates proficiency in handling these concept-related instances, indicating its capability to deal with open-world knowledge.

\begin{figure*}[tb]
\begin{center}
\includegraphics[width=0.48\linewidth]{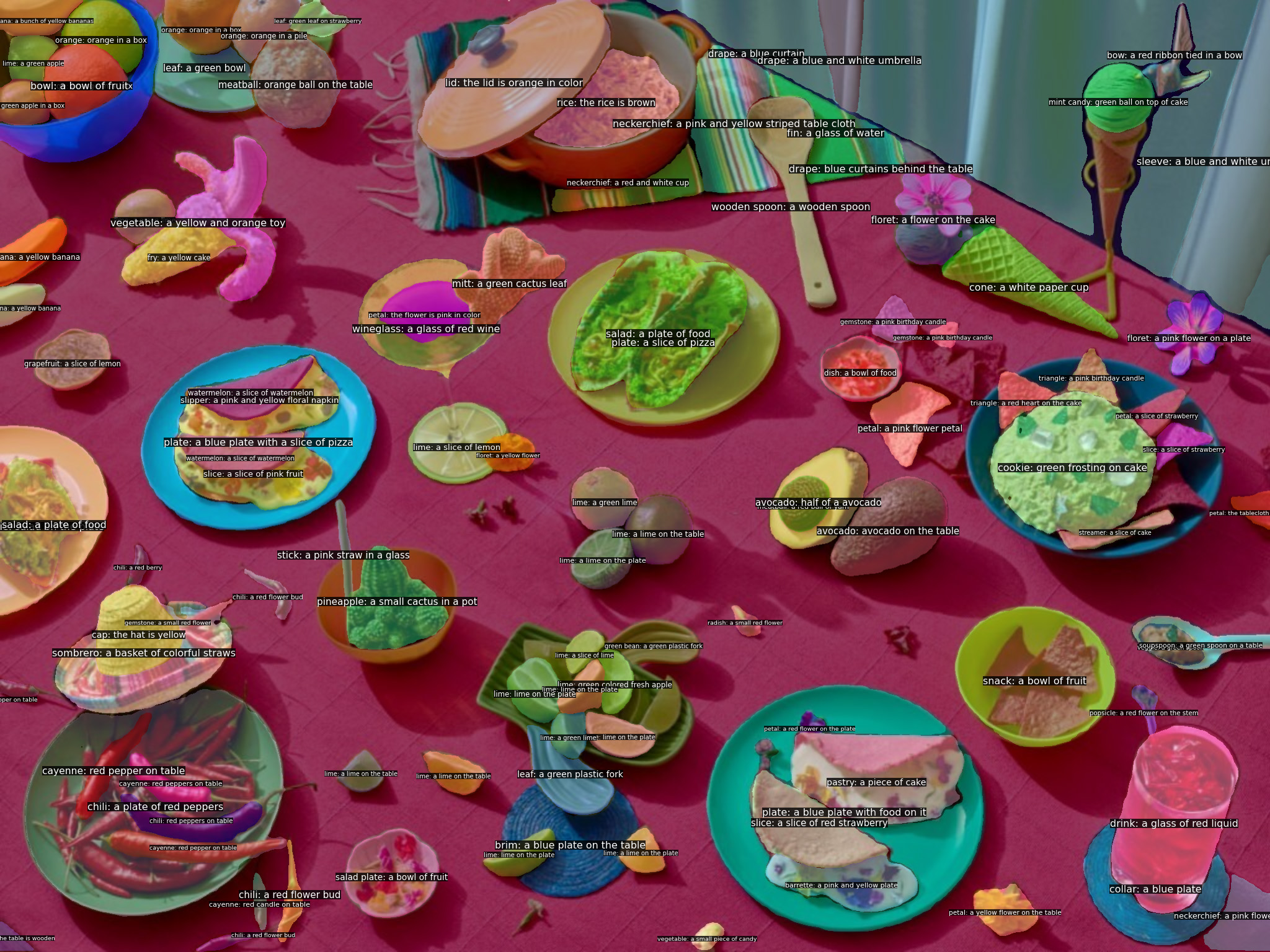}
\includegraphics[width=0.48\linewidth]{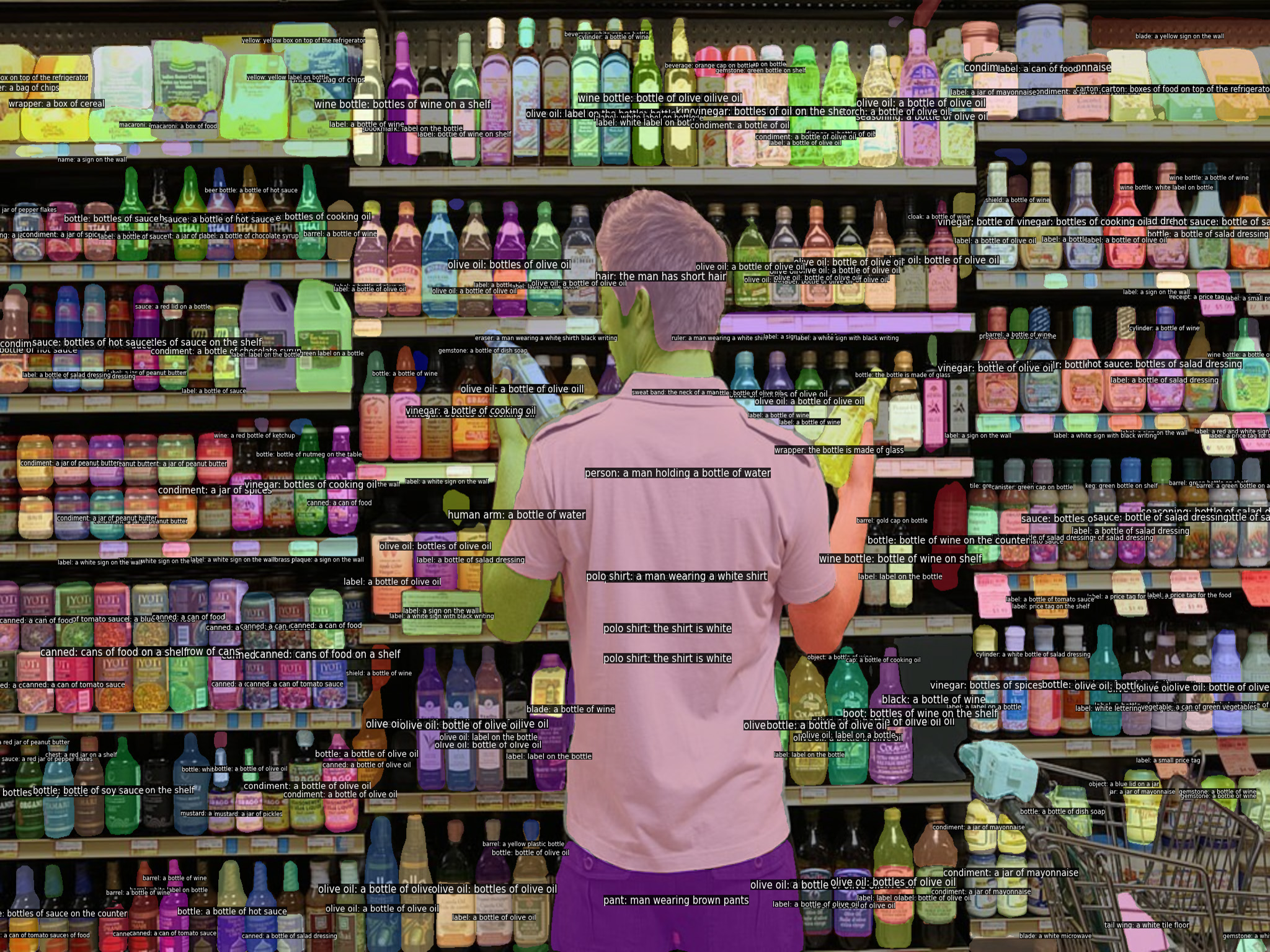}
\end{center}
\caption{\textbf{Visualization of crowd understanding.} Best viewed in color with zoom.}
\label{fig:demo_2}
\end{figure*}

\paragraph{\textbf{Crowd understanding.}} \cref{fig:demo_2} visualizes the crowd regions. \Ours accurately identifies and segments various elements within crowded or bustling environments. The segmentation masks precisely outline the distinct regions occupied by people, food as well as various uncommon commodities and stationeries. Furthermore, the accompanying caption provides an overall summary.

%% file: arxiv_v2/5_conclusion.tex
\section{Conclusion}

We propose \Ours, a promptable model for concurrently segmenting, recognizing, and captioning objects within arbitrary regions. To build such a foundation model, we explore a systematic solution that includes 1) a new dataset: injecting semantic priors from LAION-2B into SA-1B, 2) a novel framework: promptable tokenization, and 3) an effective learning method: concept prediction.  
Our key findings include:
a) Visual prompting can facilitate a broader range of tasks beyond mere segmentation.
b) SAM can be augmented with regional semantic awareness using an image-level CLIP, without compromising mask AP.
c) An orthogonal space, such as the vocabulary concept space, is essential for effective learning of CLIP priors.
TAP aims to advance segmenting anything towards perceiving anything via prompting.
We hope this work could inspire the community to develop more compact and significant vision foundation models.

%% file: arxiv_v2/arxiv_appendix.tex
\setlength{\parskip}{-5pt}
\appendix
\section*{Appendix}
\label{appendix}
This appendix consists of six parts: technical implementations of multimodal data preprocessing (Sec.~\ref{prep}), caption fine-tuning details (Sec.~\ref{caption}), additional evaluation with point prompts (Sec.~\ref{epoint}), more visualizations (Sec.~\ref{vis_f}), additional comparisons with counterparts (Sec.~\ref{comp}), and limitations (Sec.~\ref{limits}) .

\section{Pre-processing Details}\label{prep}
\begin{figure}[ht]
\centering
\includegraphics[width=0.96\linewidth]{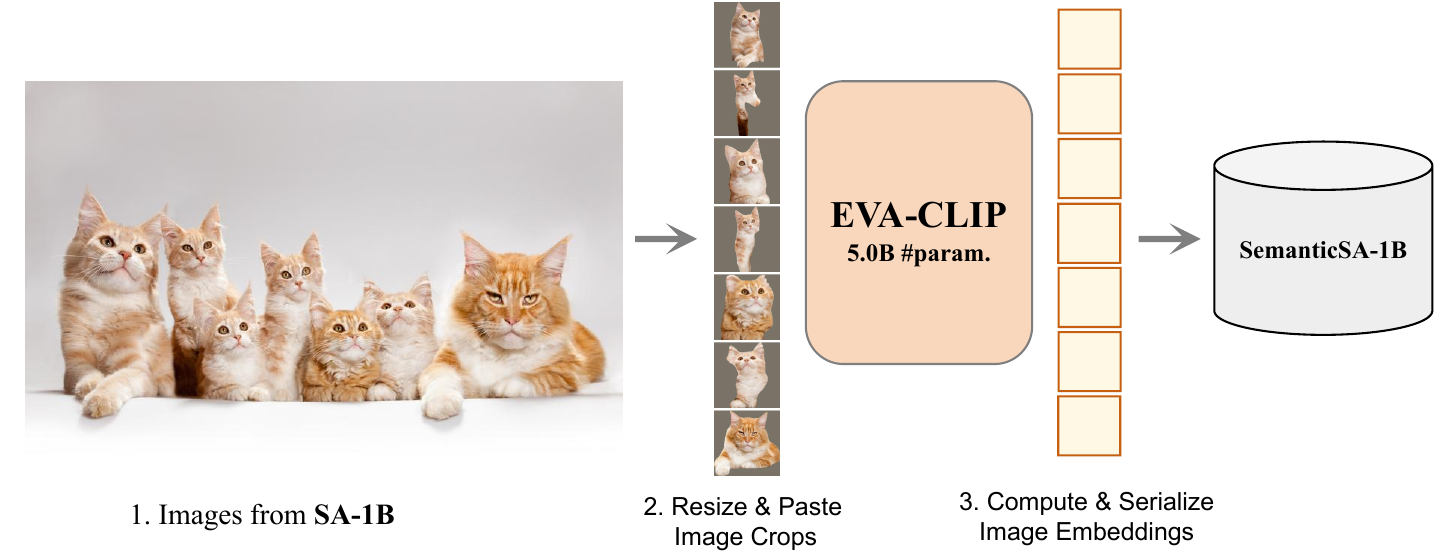}

\caption{\textbf{Pipeline for constructing SemanticSA-1B.}}
\label{fig:preprocess}
\end{figure}

\paragraph{\textbf{Vision}.} For each image in SA-1B, we crop its masked segments. Each segment is resized and then pasted onto a 224$\times$224 empty canvas, forming an `image crop'. Subsequently, we employ EVA-CLIP to compute visual embeddings $V_{C}$ for each of these crops, resulting in a total of 1.1B visual embeddings, each with a dimension of 1024 (see \cref{fig:preprocess}). These embeddings are stored in a key-value database (e.g., TFRecord), requiring approximately 2.25 TB of storage, and synchronized with SA-1B database, which is around 10.55 TB in size.

\begin{algorithm}[ht]
\caption{Pseudocode of generating text embeddings $T_{C}$.}
\lstset{basicstyle=\footnotesize}
\label{alg:vocabmaker}
\begin{python}
COCO_CLASS_NAMES = [...]
LVIS_CLASS_NAMES = [...]
ADE20K_CLASS_NAMES = [...]
OBJECTS365_CLASS_NAMES = [...]
OPEN_IMAGES_V4_BOXABLE_CLASS_NAMES = [...]
VISUAL_GENOME_CLASS_NAMES = [...]  #1600 common label strings in VG.
concepts = (COCO_CLASS_NAMES+LVIS_CLASS_NAMES+
            ADE20K_CLASS_NAMES+OBJECTS365_CLASS_NAMES+
            OPEN_IMAGES_V4_BOXABLE_CLASS_NAMES+
            VISUAL_GENOME_CLASS_NAMES)
concepts = set([name.lower() for name in concepts]) #2560 concepts in total.
remove = set()
for singular in concepts:
  for plural in [singular + 's', singular + 'es']:
    if plural in concepts:
      remove.add(plural)
concepts = sorted(list(concepts.difference(remove)))
text_templates, text_embeds = ["a {}"], []
for concept in concepts:
  texts = [tpl.format(concept) for tpl in text_templates]
  embeds = CLIP.encode_text(CLIP.tokenize(texts))
  text_embeds.append(normalize(embeds[0], dim=-1))
text_embeds = stack(text_embeds, dim=-1) * 100.
\end{python}
\end{algorithm}

\paragraph{\textbf{Language}.} We integrate concepts from prevalent image datasets, resulting in 2560 concepts. A simple prompt template, such as "a $\left\{\right\}$", is used to generate CLIP text embeddings $T_{C}$. The pseudocode for generating text embeddings is outlined in \cref{alg:vocabmaker}. These text embeddings are further used to compute the concept distribution $P_{\text{pred}}$ or $P_{\text{target}}$.

\vspace{-2pt}
\section{Caption Fine-tuning Details}\label{caption}
\paragraph{\textbf{Partial fine-tuning.}} To prevent the model from learning specific segmentation biases from the limited region caption data, we adopt a partial fine-tuning strategy after pre-training with joint masks and semantics. Specifically, we freeze the image encoder-decoder and focus solely on training the text decoder on Visual Genome (VG) data. This approach allows us to evaluate the effectiveness of the semantic token. Consequently, in this setting, promptable captioning does not contribute to segmentation.

\paragraph{\textbf{Full Fine-tuning.}} To enable end-to-end training with aligned masks, CLIP priors, and region captions, we generate masks for Visual Genome (VG) data using our partially fine-tuned model and obtain their CLIP priors using EVA-CLIP. By simultaneously enabling localization, recognition, and captioning in this setting, we observe a significant improvement in caption performance (e.g., CIDEr: partial vs. full fine-tuning: 154.7 vs. 164.7). 

\paragraph{\textbf{Next Exploration.}} The surprising results from end-to-end training suggest that our model can serve as a data engine to generate high-quality regional vision-language annotations. It is a significant asset for training large vision-language models. We leave this potential exploration for future research.

\section{Evaluation with point prompts}\label{epoint}
In the manuscript, we assess classification and caption performance using GT boxes to \textit{fairly} compare with existing methods that lack point prompting. We further evaluate our model with point prompts. Specifically, we sample $k$ loose points from the GT mask, where $k$$=$$\{1, 3, 5, 9\}$, and observe that the 5-point prompt performs comparably to the box prompt (e.g., 1-/3-/5-/9-point/box with 48.1/57.4/58.7/58.9/59.1 $AP_{\text{cls}}$ on LVIS).

\section{More Visualizations}\label{vis_f}

\begin{figure*}[ht]
\begin{center}
\includegraphics[width=0.48\linewidth]{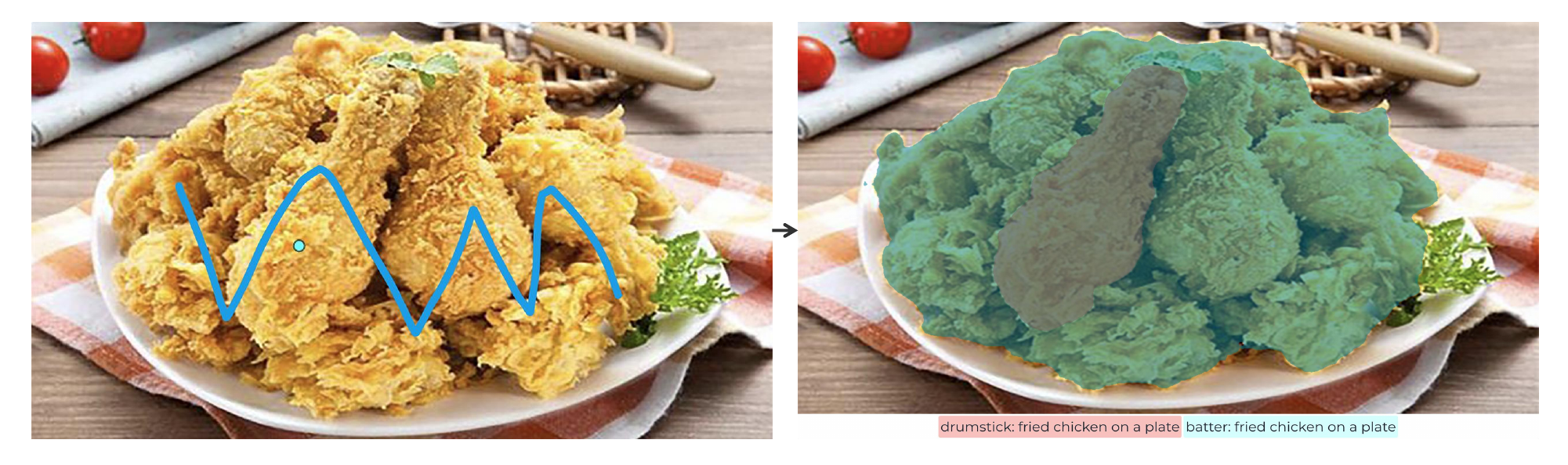}\hspace{-5pt}
\includegraphics[width=0.48\linewidth]{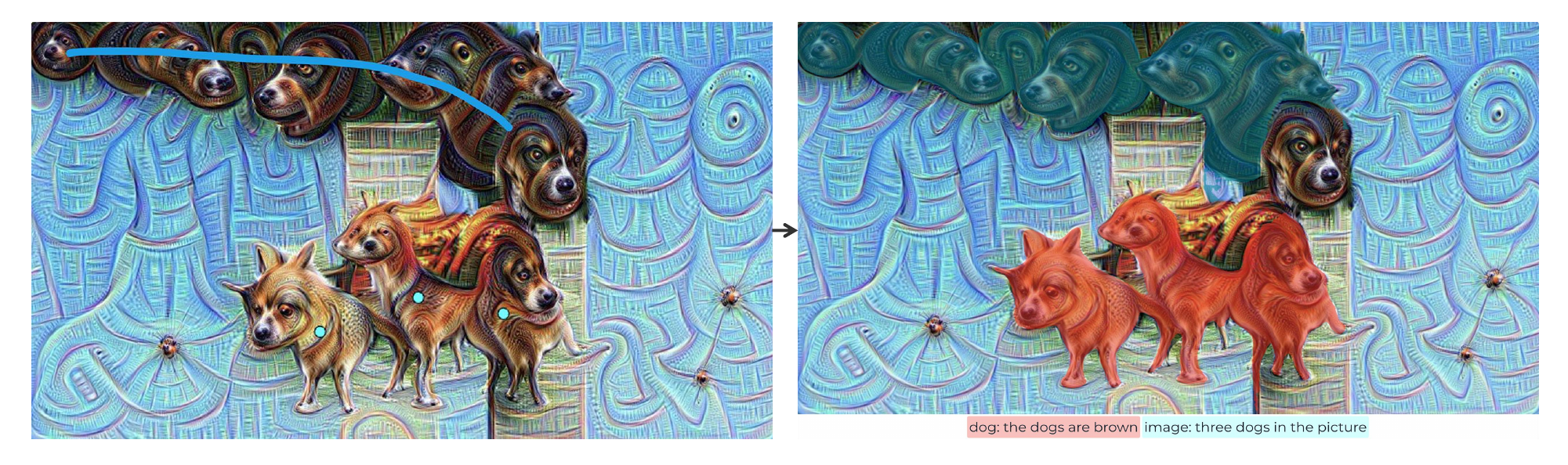}
\includegraphics[width=0.48\linewidth]{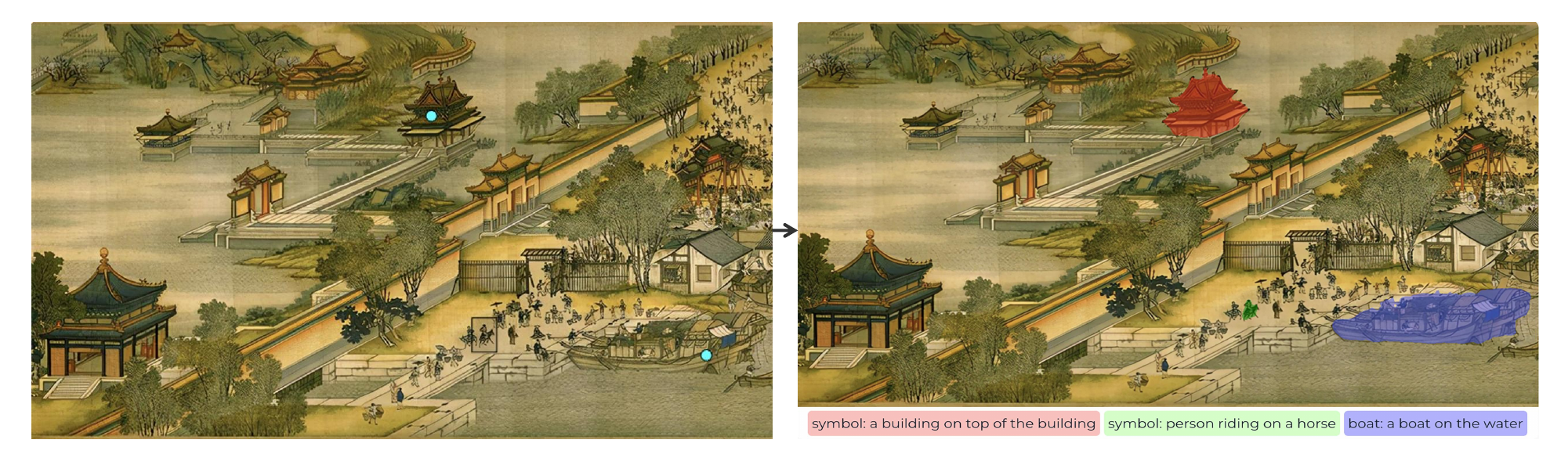}\hspace{-5pt}
\includegraphics[width=0.48\linewidth]{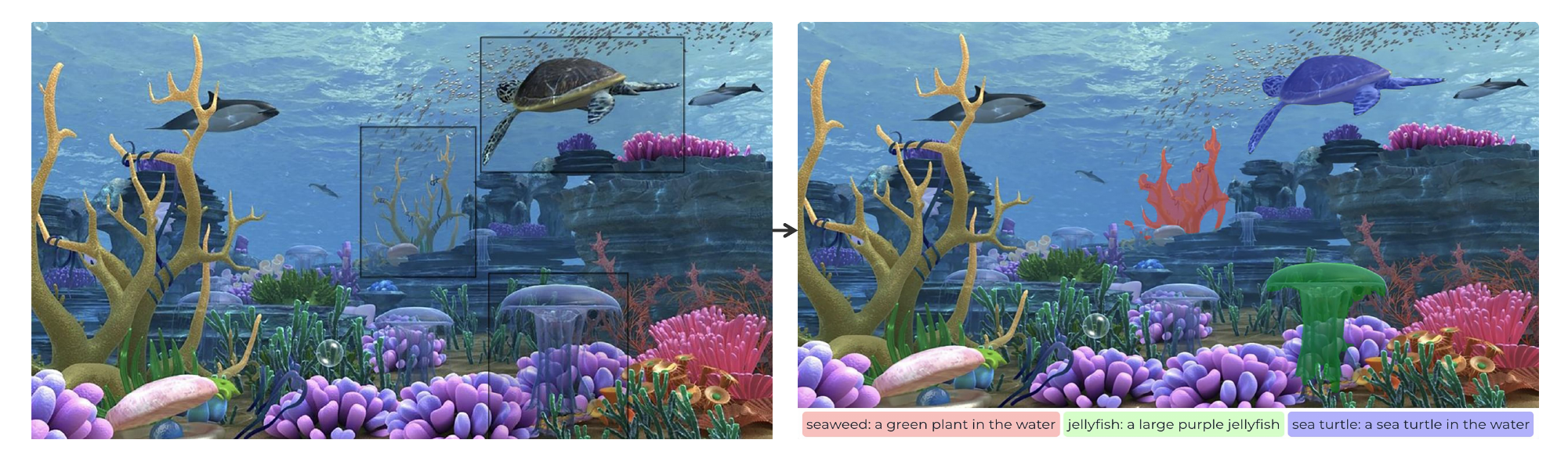}
\includegraphics[width=0.46\linewidth]{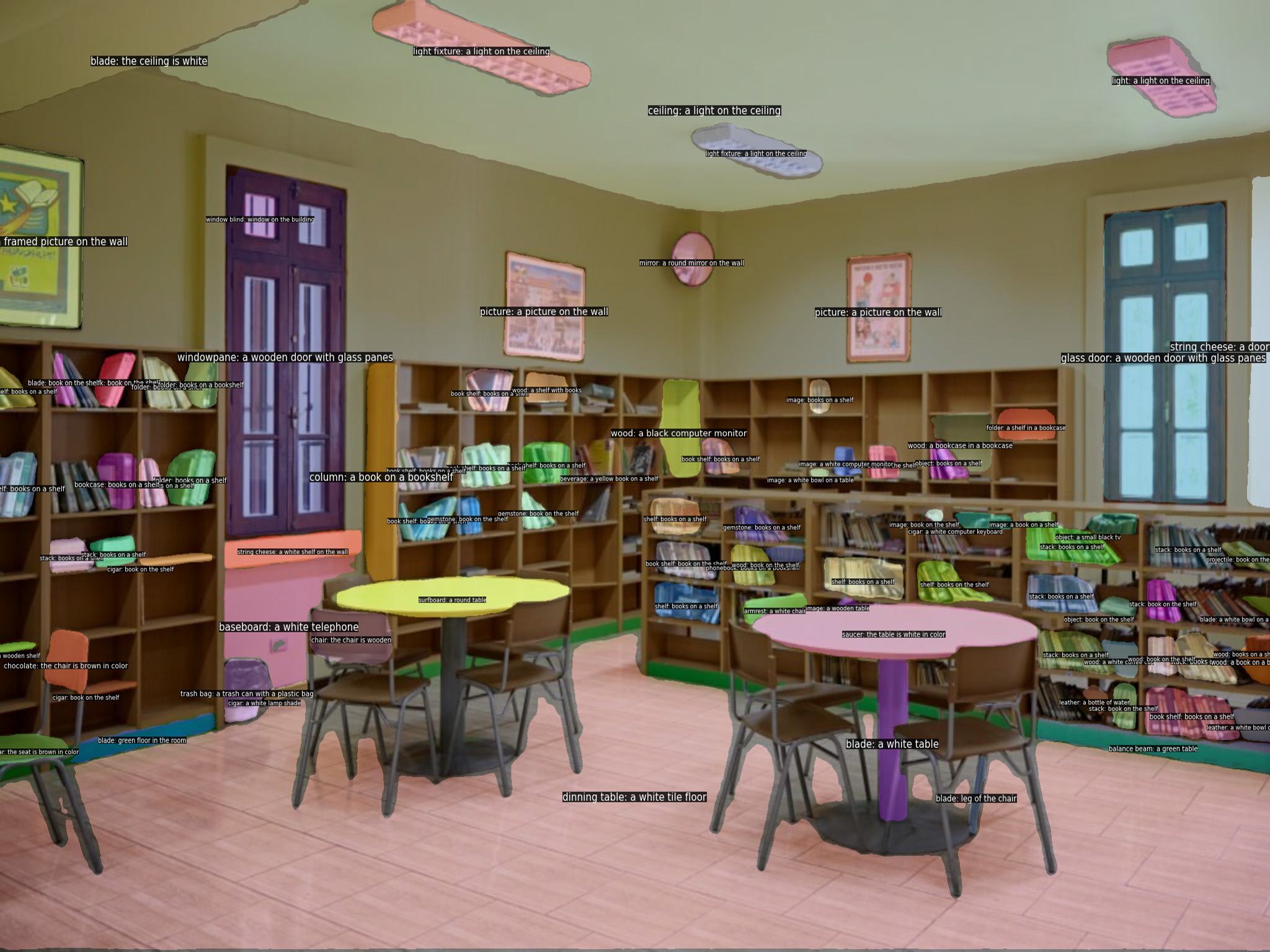}\hspace{2pt}
\includegraphics[width=0.46\linewidth]{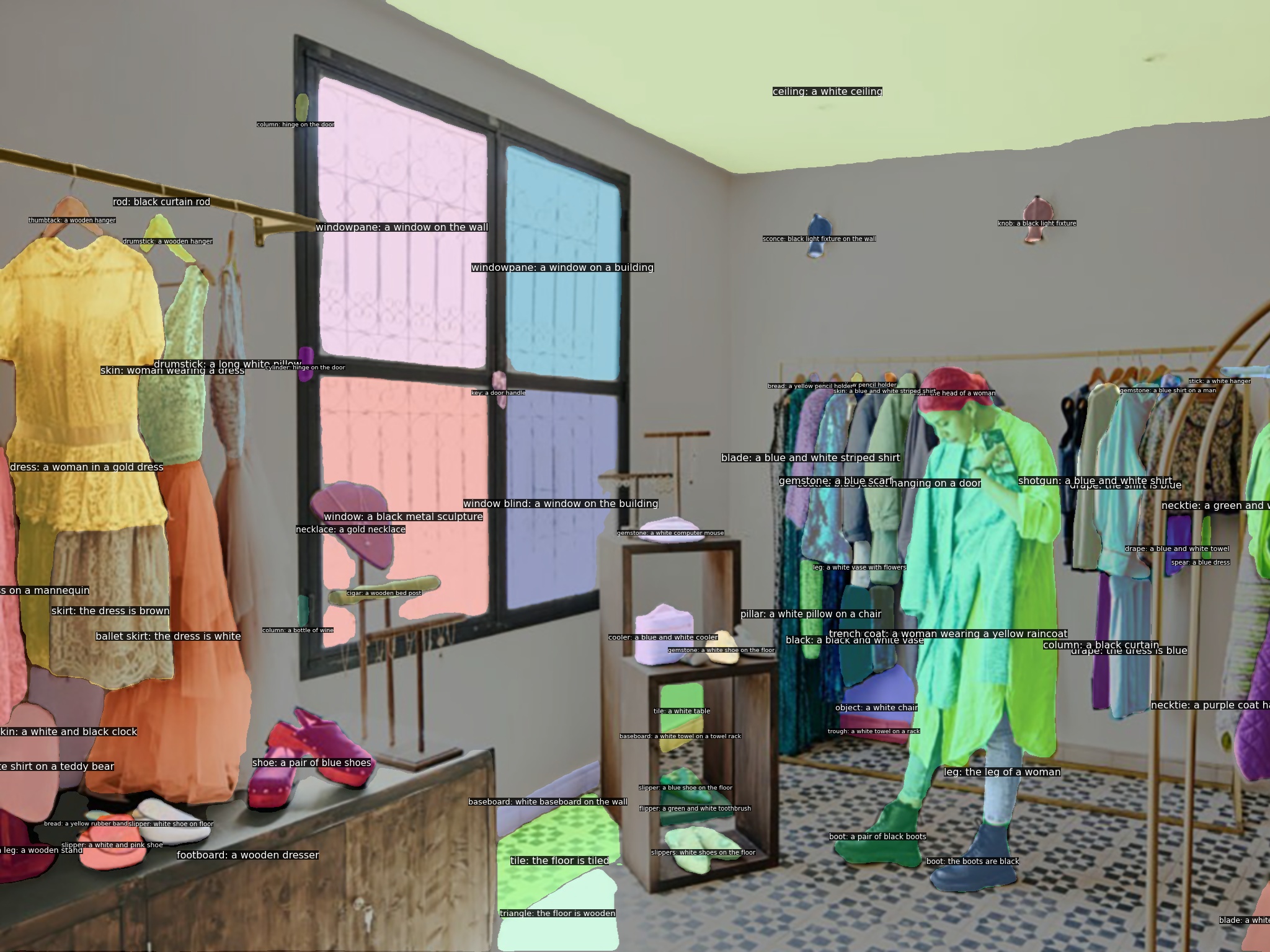}\vspace{5pt}
\includegraphics[width=0.46\linewidth]{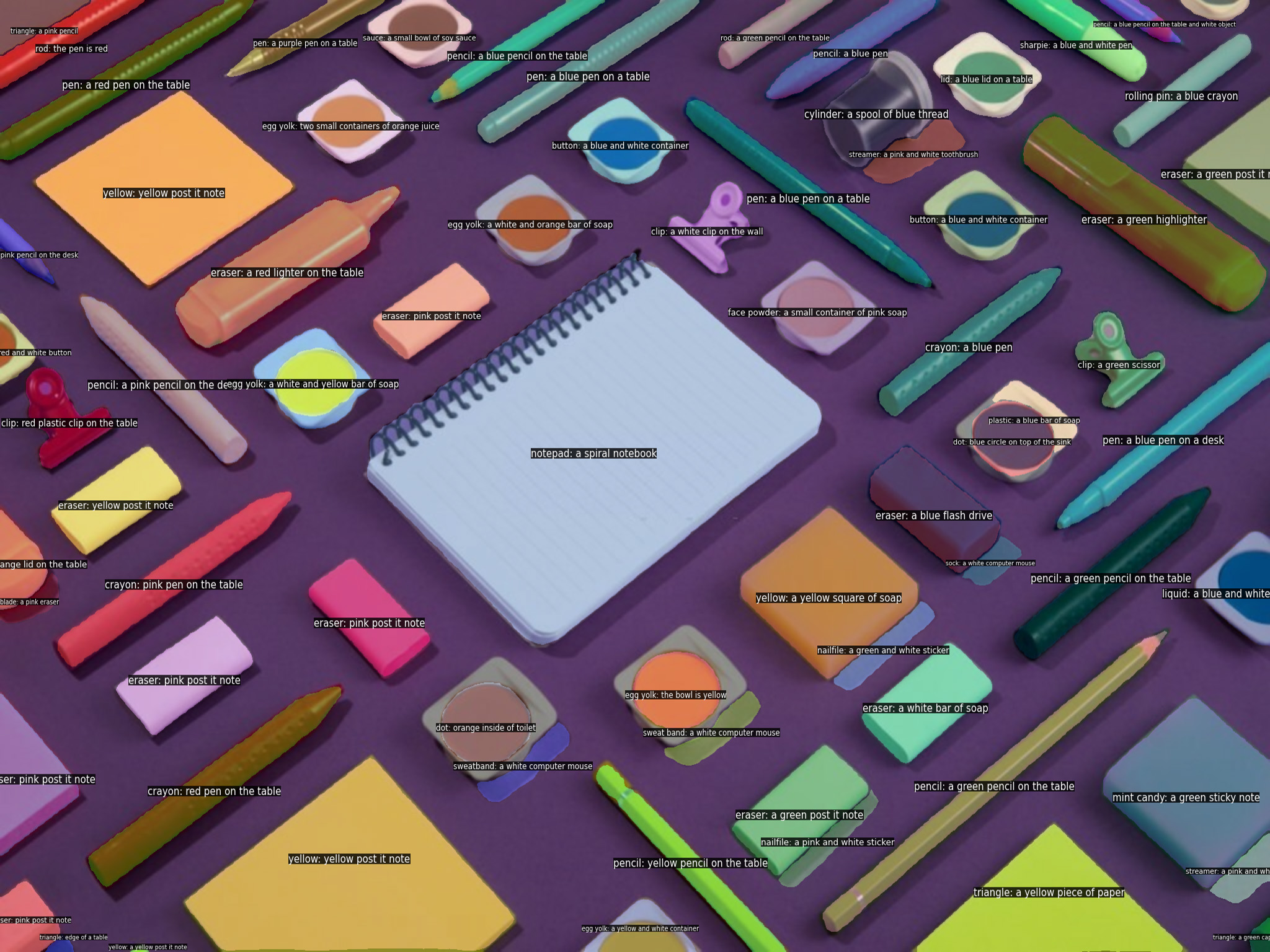}\hspace{2pt}
\includegraphics[width=0.46\linewidth]{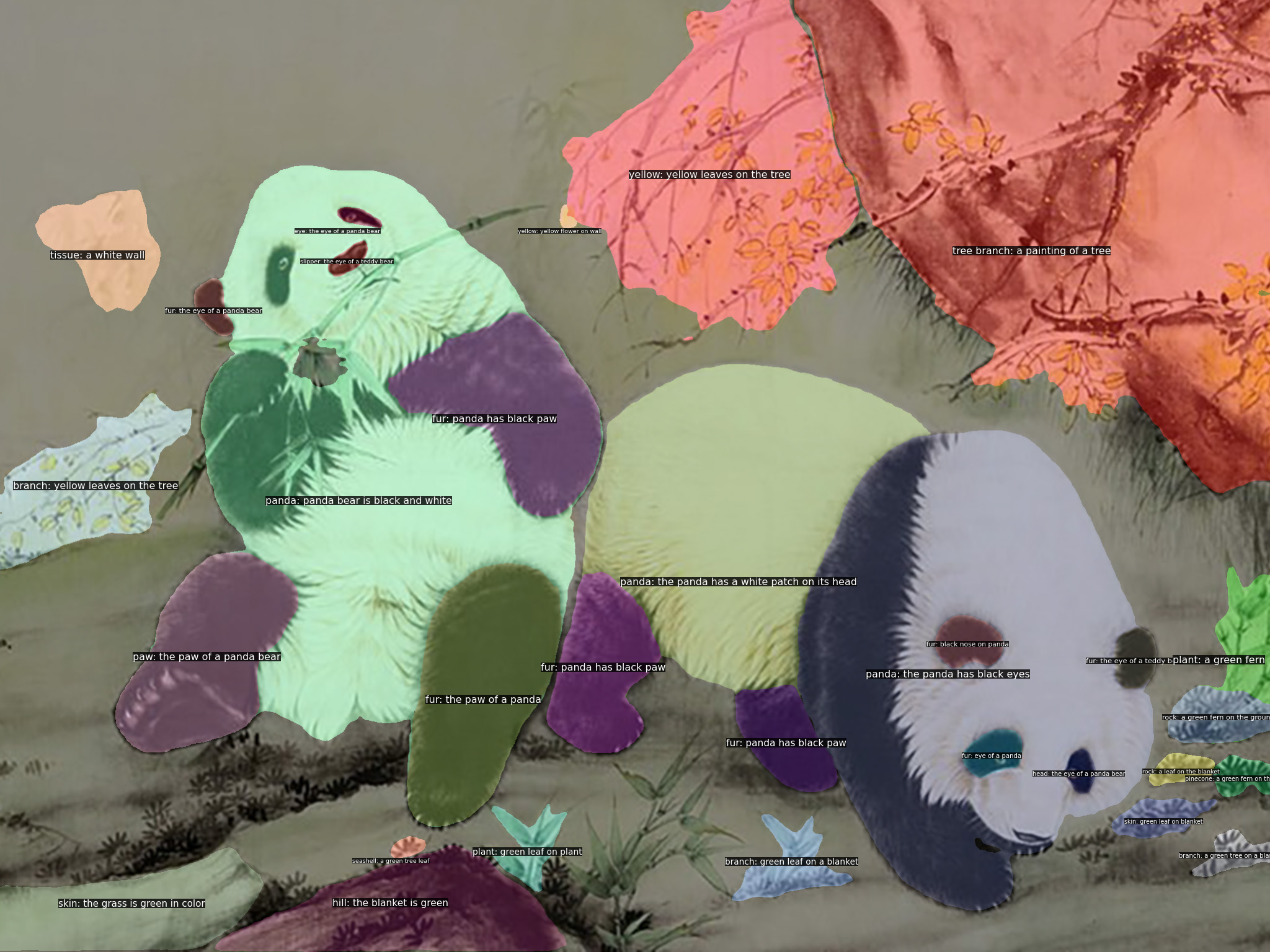}
\end{center}
\caption{\textbf{More visualizations. Best viewed in color with zoom.}}
\label{fig:demo_4}
\end{figure*}

\cref{fig:demo_4} provides additional visualizations using various prompts. As observed, our model is able to perform accurate segmentation, recognition, and captioning of objects, demonstrating the proficiency even with artistic images.

\begin{figure}[ht!]
\begin{center}
\includegraphics[width=0.96\linewidth]{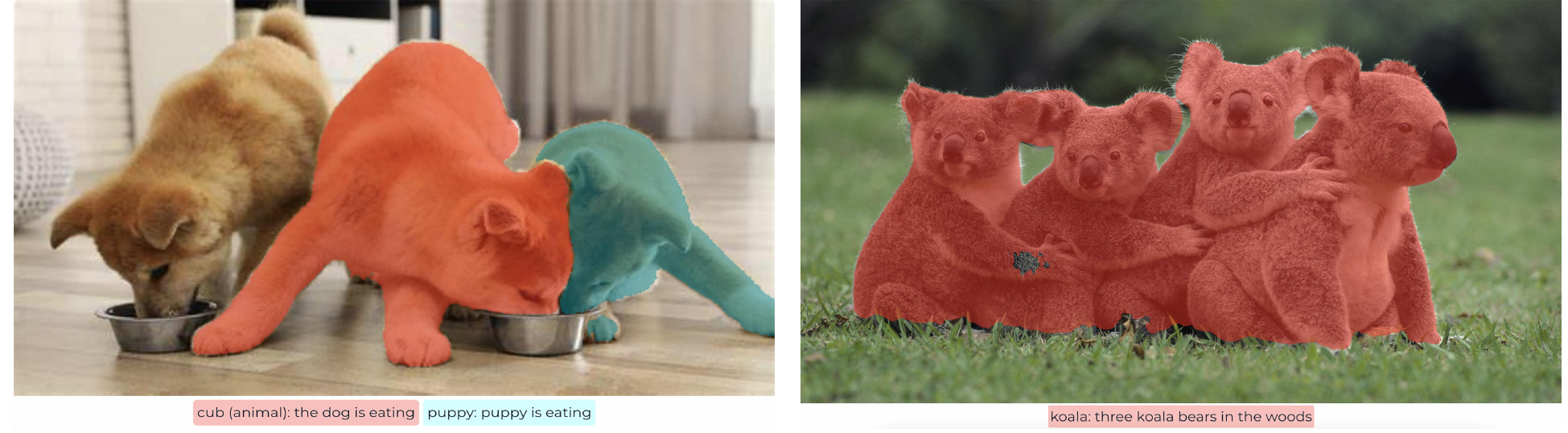}
\end{center}
\caption{\textbf{Visualization of failure cases.}}
\label{fig:supp_f}
\end{figure}

\section{Comparison with SEEM and SAM-CLIP}\label{comp}
All three models aim to find a joint visual-semantic space. SEEM~\cite{seem}, trained on COCO and LVIS,  predicts object queries and semantic masks. Conversely, our TAP predicts semantic masks for SA-1B regions. Due to lack of object queries, we prompt TAP with ViTDet boxes. In this setting: predicting masks from queries and predicting categories from CLIP text embeddings, TAP marginally outperforms SEEM in $\textit{class-specific}$ instance segmentation tasks (\cref{tab:morecomp}). On the other hand, SAM-CLIP~\cite{samclip} distills CLIP and SAM features into two task heads for semantic segmentation (\eg, segmentation in ADE20K) and \textit{class-agnostic} instance segmentation. Although each head retains its original functionality, integrating them for \textit{class-specific} instance segmentation is not straightforward. In contrast, TAP naturally performs this segmentation task and achieves better results (\cref{tab:morecomp}).

\begin{table*}[h]
\vspace{-1em}
\caption{\textbf{Zero-shot Instance \& Semantic Segmentation.} InstSeg: box prompts
from ViTDet; SemSeg: point prompts with a 16 $\times$ 16 grid.}\label{tab:morecomp}
\centering
\setlength{\tabcolsep}{4mm}
\renewcommand\arraystretch{1}
\resizebox{0.8\linewidth}{!}{

\footnotesize
\begin{tabular}{c|cc|c}
\toprule
Model & COCO$_{\text{inst}}$ & LVIS$_{\text{inst}}$ & ADE20K$_{\text{sem}}$ \\
\midrule
SAM-CLIP-B \scriptsize {[52]}     & 40.9 & 35.0 & 17.1    \\
TAP-B  & 45.1 & 42.2 & 17.8   \\
% \midrule
SEEM-L \scriptsize {[78]}    & 47.7 & - & - \\
TAP-L  & 48.2 & 43.2 & 19.9 \\
\bottomrule  
\end{tabular} }
\vspace{-25pt}
\end{table*}

\section{Limitations}\label{limits}
Despite its advancements, \Ours has two main constraints. It is trained using a human-curated label space, which still falls short of an open-world assumption. This constraint also leads to an unstable ranking of similar concepts during inference (\cref{fig:supp_f} left). Additionally, the text decoder, fine-tuned on a constrained set of region caption data, may limit the model’s scalability and breadth of vision-language understanding. For example, object counting cannot be solved well with simple annotations on the quantity (\cref{fig:supp_f} right). Expanding the caption data is expected to instruct models for complex understanding.

%% file: main.bbl
\begin{thebibliography}{10}
\providecommand{\url}[1]{\texttt{#1}}
\providecommand{\urlprefix}{URL }
\providecommand{\doi}[1]{https://doi.org/#1}

\bibitem{alayrac2022flamingo}
Alayrac, J.B., Donahue, J., Luc, P., Miech, A., Barr, I., Hasson, Y., Lenc, K., Mensch, A., Millican, K., Reynolds, M., et~al.: Flamingo: a visual language model for few-shot learning. Advances in Neural Information Processing Systems  \textbf{35},  23716--23736 (2022)

\bibitem{bolya2019yolact}
Bolya, D., Zhou, C., Xiao, F., Lee, Y.J.: Yolact: Real-time instance segmentation. In: ICCV. pp. 9157--9166 (2019)

\bibitem{changpinyo2021cc12m}
Changpinyo, S., Sharma, P., Ding, N., Soricut, R.: {Conceptual 12M}: Pushing web-scale image-text pre-training to recognize long-tail visual concepts. In: CVPR (2021)

\bibitem{maskformer}
Cheng, B., Schwing, A., Kirillov, A.: Per-pixel classification is not all you need for semantic segmentation. NeurIPS  \textbf{34},  17864--17875 (2021)

\bibitem{zegformer}
Ding, J., Xue, N., Xia, G.S., Dai, D.: Decoupling zero-shot semantic segmentation. In: CVPR. pp. 11583--11592 (2022)

\bibitem{MaskCLIP}
Ding, Z., Wang, J., Tu, Z.: Open-vocabulary panoptic segmentation with maskclip. arXiv preprint arXiv:2208.08984  (2022)

\bibitem{dosovitskiy2020image}
Dosovitskiy, A., Beyer, L., Kolesnikov, A., Weissenborn, D., Zhai, X., Unterthiner, T., Dehghani, M., Minderer, M., Heigold, G., Gelly, S., et~al.: An image is worth 16x16 words: Transformers for image recognition at scale. In: ICLR (2020)

\bibitem{geng2023openllama}
Geng, X., Liu, H.: Openllama: An open reproduction of llama. URL: https://github. com/openlm-research/open\_llama  (2023)

\bibitem{ghiasi2021simple}
Ghiasi, G., Cui, Y., Srinivas, A., Qian, R., Lin, T.Y., Cubuk, E.D., Le, Q.V., Zoph, B.: Simple copy-paste is a strong data augmentation method for instance segmentation. In: CVPR. pp. 2918--2928 (2021)

\bibitem{OpenSeg}
Ghiasi, G., Gu, X., Cui, Y., Lin, T.Y.: Scaling open-vocabulary image segmentation with image-level labels. In: ECCV. pp. 540--557. Springer (2022)

\bibitem{gu2021open}
Gu, X., Lin, T.Y., Kuo, W., Cui, Y.: Open-vocabulary detection via vision and language knowledge distillation. arXiv preprint arXiv:2104.13921  (2021)

\bibitem{gupta2019lvis}
Gupta, A., Dollar, P., Girshick, R.: Lvis: A dataset for large vocabulary instance segmentation. In: CVPR. pp. 5356--5364 (2019)

\bibitem{he2022masked}
He, K., Chen, X., Xie, S., Li, Y., Doll{\'a}r, P., Girshick, R.: Masked autoencoders are scalable vision learners. In: CVPR. pp. 16000--16009 (2022)

\bibitem{he2017mask}
He, K., Gkioxari, G., Doll{\'a}r, P., Girshick, R.: Mask r-cnn. In: ICCV. pp. 2961--2969 (2017)

\bibitem{huang2016deep}
Huang, G., Sun, Y., Liu, Z., Sedra, D., Weinberger, K.Q.: Deep networks with stochastic depth. In: ECCV. pp. 646--661. Springer (2016)

\bibitem{XPM}
Huynh, D., Kuen, J., Lin, Z., Gu, J., Elhamifar, E.: Open-vocabulary instance segmentation via robust cross-modal pseudo-labeling. In: CVPR. pp. 7020--7031 (2022)

\bibitem{jacobs1991adaptive}
Jacobs, R.A., Jordan, M.I., Nowlan, S.J., Hinton, G.E.: Adaptive mixtures of local experts. Neural computation  \textbf{3}(1),  79--87 (1991)

\bibitem{jia2021scaling}
Jia, C., Yang, Y., Xia, Y., Chen, Y.T., Parekh, Z., Pham, H., Le, Q., Sung, Y.H., Li, Z., Duerig, T.: Scaling up visual and vision-language representation learning with noisy text supervision. In: International conference on machine learning. pp. 4904--4916. PMLR (2021)

\bibitem{OVDiff}
Karazija, L., Laina, I., Vedaldi, A., Rupprecht, C.: Diffusion models for zero-shot open-vocabulary segmentation. arXiv preprint arXiv:2306.09316  (2023)

\bibitem{sam}
Kirillov, A., Mintun, E., Ravi, N., Mao, H., Rolland, C., Gustafson, L., Xiao, T., Whitehead, S., Berg, A.C., Lo, W.Y., et~al.: Segment anything. arXiv preprint arXiv:2304.02643  (2023)

\bibitem{krishna2017visual}
Krishna, R., Zhu, Y., Groth, O., Johnson, J., Hata, K., Kravitz, J., Chen, S., Kalantidis, Y., Li, L.J., Shamma, D.A., et~al.: Visual genome: Connecting language and vision using crowdsourced dense image annotations. IJCV  \textbf{123},  32--73 (2017)

\bibitem{kuo2022f}
Kuo, W., Cui, Y., Gu, X., Piergiovanni, A., Angelova, A.: F-vlm: Open-vocabulary object detection upon frozen vision and language models. arXiv preprint arXiv:2209.15639  (2022)

\bibitem{kuznetsova2020open}
Kuznetsova, A., Rom, H., Alldrin, N., Uijlings, J., Krasin, I., Pont-Tuset, J., Kamali, S., Popov, S., Malloci, M., Kolesnikov, A., et~al.: The open images dataset v4: Unified image classification, object detection, and visual relationship detection at scale. IJCV  \textbf{128}(7),  1956--1981 (2020)

\bibitem{LSeg}
Li, B., Weinberger, K.Q., Belongie, S., Koltun, V., Ranftl, R.: Language-driven semantic segmentation (2022)

\bibitem{semantic-sam}
Li, F., Zhang, H., Sun, P., Zou, X., Liu, S., Yang, J., Li, C., Zhang, L., Gao, J.: Semantic-sam: Segment and recognize anything at any granularity. arXiv preprint arXiv:2307.04767  (2023)

\bibitem{li2023scaling}
Li, Y., Fan, H., Hu, R., Feichtenhofer, C., He, K.: Scaling language-image pre-training via masking. In: Proceedings of the IEEE/CVF Conference on Computer Vision and Pattern Recognition. pp. 23390--23400 (2023)

\bibitem{li2022exploring}
Li, Y., Mao, H., Girshick, R., He, K.: Exploring plain vision transformer backbones for object detection. In: ECCV. pp. 280--296. Springer (2022)

\bibitem{li2022mvitv2}
Li, Y., Wu, C.Y., Fan, H., Mangalam, K., Xiong, B., Malik, J., Feichtenhofer, C.: Mvitv2: Improved multiscale vision transformers for classification and detection. In: CVPR. pp. 4804--4814 (2022)

\bibitem{OVSeg}
Liang, F., Wu, B., Dai, X., Li, K., Zhao, Y., Zhang, H., Zhang, P., Vajda, P., Marculescu, D.: Open-vocabulary semantic segmentation with mask-adapted clip. In: CVPR. pp. 7061--7070 (2023)

\bibitem{lin2017focal}
Lin, T.Y., Goyal, P., Girshick, R., He, K., Doll{\'a}r, P.: Focal loss for dense object detection. In: ICCV. pp. 2980--2988 (2017)

\bibitem{coco}
Lin, T.Y., Maire, M., Belongie, S., Hays, J., Perona, P., Ramanan, D., Doll{\'a}r, P., Zitnick, C.L.: Microsoft coco: Common objects in context. In: ECCV. pp. 740--755. Springer (2014)

\bibitem{liu2023internchat}
Liu, Z., He, Y., Wang, W., Wang, W., Wang, Y., Chen, S., Zhang, Q., Yang, Y., Li, Q., Yu, J., et~al.: Internchat: Solving vision-centric tasks by interacting with chatbots beyond language. arXiv preprint arXiv:2305.05662  (2023)

\bibitem{long2015fully}
Long, J., Shelhamer, E., Darrell, T.: Fully convolutional networks for semantic segmentation. In: CVPR. pp. 3431--3440 (2015)

\bibitem{loshchilov2017sgdr}
Loshchilov, I., Hutter, F.: Sgdr: Stochastic gradient descent with warm restarts. In: ICLR (2017)

\bibitem{loshchilov2019decoupled}
Loshchilov, I., Hutter, F.: Decoupled weight decay regularization. In: ICLR (2019)

\bibitem{lu2022unified}
Lu, J., Clark, C., Zellers, R., Mottaghi, R., Kembhavi, A.: Unified-io: A unified model for vision, language, and multi-modal tasks. arXiv preprint arXiv:2206.08916  (2022)

\bibitem{mao2016generation}
Mao, J., Huang, J., Toshev, A., Camburu, O., Yuille, A.L., Murphy, K.: Generation and comprehension of unambiguous object descriptions. In: CVPR. pp. 11--20 (2016)

\bibitem{milletari2016fully}
Milletari, F., Navab, N., Ahmadi, S., Net, V.: Fully convolutional neural networks for volumetric medical image segmentation. In: Proceedings of the 2016 Fourth International Conference on 3D Vision (3DV). pp. 565--571 (2016)

\bibitem{minderer2023scaling}
Minderer, M., Gritsenko, A., Houlsby, N.: Scaling open-vocabulary object detection. arXiv preprint arXiv:2306.09683  (2023)

\bibitem{FreeSeg}
Qin, J., Wu, J., Yan, P., Li, M., Yuxi, R., Xiao, X., Wang, Y., Wang, R., Wen, S., Pan, X., et~al.: Freeseg: Unified, universal and open-vocabulary image segmentation. In: CVPR. pp. 19446--19455 (2023)

\bibitem{radford2021learning}
Radford, A., Kim, J.W., Hallacy, C., Ramesh, A., Goh, G., Agarwal, S., Sastry, G., Askell, A., Mishkin, P., Clark, J., et~al.: Learning transferable visual models from natural language supervision. In: ICML. pp. 8748--8763 (2021)

\bibitem{schuhmann2022laion}
Schuhmann, C., Beaumont, R., Vencu, R., Gordon, C., Wightman, R., Cherti, M., Coombes, T., Katta, A., Mullis, C., Wortsman, M., et~al.: Laion-5b: An open large-scale dataset for training next generation image-text models. arXiv preprint arXiv:2210.08402  (2022)

\bibitem{sennrich2015neural}
Sennrich, R., Haddow, B., Birch, A.: Neural machine translation of rare words with subword units. arXiv preprint arXiv:1508.07909  (2015)

\bibitem{shao2019objects365}
Shao, S., Li, Z., Zhang, T., Peng, C., Yu, G., Zhang, X., Li, J., Sun, J.: Objects365: A large-scale, high-quality dataset for object detection. In: ICCV. pp. 8430--8439 (2019)

\bibitem{sharma2018conceptual}
Sharma, P., Ding, N., Goodman, S., Soricut, R.: Conceptual captions: A cleaned, hypernymed, image alt-text dataset for automatic image captioning. In: ACL. pp. 2556--2565 (2018)

\bibitem{srivastava2014dropout}
Srivastava, N., Hinton, G., Krizhevsky, A., Sutskever, I., Salakhutdinov, R.: Dropout: a simple way to prevent neural networks from overfitting. The journal of machine learning research  \textbf{15}(1),  1929--1958 (2014)

\bibitem{su2021roformer}
Su, J., Lu, Y., Pan, S., Murtadha, A., Wen, B., Liu, Y.: Roformer: Enhanced transformer with rotary position embedding. arXiv preprint arXiv:2104.09864  (2021)

\bibitem{sun2023evaclip}
Sun, Q., Fang, Y., Wu, L., Wang, X., Cao, Y.: Eva-clip: Improved training techniques for clip at scale. arXiv preprint arXiv:2303.15389  (2023)

\bibitem{sun2023generative}
Sun, Q., Yu, Q., Cui, Y., Zhang, F., Zhang, X., Wang, Y., Gao, H., Liu, J., Huang, T., Wang, X.: Generative pretraining in multimodality. arXiv preprint arXiv:2307.05222  (2023)

\bibitem{sun2024vrp}
Sun, Y., Chen, J., Zhang, S., Zhang, X., Chen, Q., Zhang, G., Ding, E., Wang, J., Li, Z.: Vrp-sam: Sam with visual reference prompt. arXiv preprint arXiv:2402.17726  (2024)

\bibitem{sun2023alpha}
Sun, Z., Fang, Y., Wu, T., Zhang, P., Zang, Y., Kong, S., Xiong, Y., Lin, D., Wang, J.: Alpha-clip: A clip model focusing on wherever you want. arXiv preprint arXiv:2312.03818  (2023)

\bibitem{samclip}
Wang, H., Vasu, P.K.A., Faghri, F., Vemulapalli, R., Farajtabar, M., Mehta, S., Rastegari, M., Tuzel, O., Pouransari, H.: Sam-clip: Merging vision foundation models towards semantic and spatial understanding. arXiv preprint arXiv:2310.15308  (2023)

\bibitem{wang2023object}
Wang, L., Liu, Y., Du, P., Ding, Z., Liao, Y., Qi, Q., Chen, B., Liu, S.: Object-aware distillation pyramid for open-vocabulary object detection. In: CVPR. pp. 11186--11196 (2023)

\bibitem{wang2023caption}
Wang, T., Zhang, J., Fei, J., Ge, Y., Zheng, H., Tang, Y., Li, Z., Gao, M., Zhao, S., Shan, Y., et~al.: Caption anything: Interactive image description with diverse multimodal controls. arXiv preprint arXiv:2305.02677  (2023)

\bibitem{asm}
Wang, W., Shi, M., Li, Q., Wang, W., Huang, Z., Xing, L., Chen, Z., Li, H., Zhu, X., Cao, Z., et~al.: The all-seeing project: Towards panoptic visual recognition and understanding of the open world. arXiv preprint arXiv:2308.01907  (2023)

\bibitem{SOLO}
Wang, X., Kong, T., Shen, C., Jiang, Y., Li, L.: Solo: Segmenting objects by locations. In: ECCV. pp. 649--665. Springer (2020)

\bibitem{Painter}
Wang, X., Wang, W., Cao, Y., Shen, C., Huang, T.: Images speak in images: A generalist painter for in-context visual learning. arXiv preprint arXiv:2212.02499  (2022)

\bibitem{SOLOV2}
Wang, X., Zhang, R., Kong, T., Li, L., Shen, C.: Solov2: Dynamic and fast instance segmentation. NeurIPS  \textbf{33},  17721--17732 (2020)

\bibitem{wang2023seggpt}
Wang, X., Zhang, X., Cao, Y., Wang, W., Shen, C., Huang, T.: Seggpt: Towards segmenting everything in context. In: ICCV. pp. 1130--1140 (2023)

\bibitem{wu2022grit}
Wu, J., Wang, J., Yang, Z., Gan, Z., Liu, Z., Yuan, J., Wang, L.: Grit: A generative region-to-text transformer for object understanding. arXiv preprint arXiv:2212.00280  (2022)

\bibitem{CGG}
Wu, J., Li, X., Ding, H., Li, X., Cheng, G., Tong, Y., Loy, C.C.: Betrayed by captions: Joint caption grounding and generation for open vocabulary instance segmentation. arXiv preprint arXiv:2301.00805  (2023)

\bibitem{xiao2018unified}
Xiao, T., Liu, Y., Zhou, B., Jiang, Y., Sun, J.: Unified perceptual parsing for scene understanding. In: European Conference on Computer Vision. Springer (2018)

\bibitem{huang2023sca}
Xiaoke, H., Jianfeng, W., Yansong, T., Zheng, Z., Han, H., Jiwen, L., Lijuan, W., Zicheng, L.: Segment and caption anything. arXiv preprint arXiv:2312.00869  (2023)

\bibitem{xiong2023efficientsam}
Xiong, Y., Varadarajan, B., Wu, L., Xiang, X., Xiao, F., Zhu, C., Dai, X., Wang, D., Sun, F., Iandola, F., et~al.: Efficientsam: Leveraged masked image pretraining for efficient segment anything. arXiv preprint arXiv:2312.00863  (2023)

\bibitem{groupvit}
Xu, J., De~Mello, S., Liu, S., Byeon, W., Breuel, T., Kautz, J., Wang, X.: Groupvit: Semantic segmentation emerges from text supervision. In: CVPR. pp. 18134--18144 (2022)

\bibitem{ODISE}
Xu, J., Liu, S., Vahdat, A., Byeon, W., Wang, X., De~Mello, S.: Open-vocabulary panoptic segmentation with text-to-image diffusion models. In: CVPR. pp. 2955--2966 (2023)

\bibitem{yang2023recognize}
Yang, H., Ma, C., Wen, B., Jiang, Y., Yuan, Z., Zhu, X.: Recognize any regions. arXiv preprint arXiv:2311.01373  (2023)

\bibitem{yao2022detclip}
Yao, L., Han, J., Wen, Y., Liang, X., Xu, D., Zhang, W., Li, Z., Xu, C., Xu, H.: Detclip: Dictionary-enriched visual-concept paralleled pre-training for open-world detection. NeurIPS  \textbf{35},  9125--9138 (2022)

\bibitem{zareian2021open}
Zareian, A., Rosa, K.D., Hu, D.H., Chang, S.F.: Open-vocabulary object detection using captions. In: CVPR. pp. 14393--14402 (2021)

\bibitem{zhang2023faster}
Zhang, C., Han, D., Qiao, Y., Kim, J.U., Bae, S.H., Lee, S., Hong, C.S.: Faster segment anything: Towards lightweight sam for mobile applications. arXiv preprint arXiv:2306.14289  (2023)

\bibitem{OpenSeeD}
Zhang, H., Li, F., Zou, X., Liu, S., Li, C., Yang, J., Zhang, L.: A simple framework for open-vocabulary segmentation and detection. In: ICCV. pp. 1020--1031 (2023)

\bibitem{zhang2023gpt4roi}
Zhang, S., Sun, P., Chen, S., Xiao, M., Shao, W., Zhang, W., Chen, K., Luo, P.: Gpt4roi: Instruction tuning large language model on region-of-interest. arXiv preprint arXiv:2307.03601  (2023)

\bibitem{zheng2023judging}
Zheng, L., Chiang, W.L., Sheng, Y., Zhuang, S., Wu, Z., Zhuang, Y., Lin, Z., Li, Z., Li, D., Xing, E., et~al.: Judging llm-as-a-judge with mt-bench and chatbot arena. arXiv preprint arXiv:2306.05685  (2023)

\bibitem{zhong2022regionclip}
Zhong, Y., Yang, J., Zhang, P., Li, C., Codella, N., Li, L.H., Zhou, L., Dai, X., Yuan, L., Li, Y., et~al.: Regionclip: Region-based language-image pretraining. In: CVPR. pp. 16793--16803 (2022)

\bibitem{zhou2017scene}
Zhou, B., Zhao, H., Puig, X., Fidler, S., Barriuso, A., Torralba, A.: Scene parsing through ade20k dataset. In: CVPR. pp. 633--641 (2017)

\bibitem{MaskCLIP+}
Zhou, C., Loy, C.C., Dai, B.: Extract free dense labels from clip. In: ECCV. pp. 696--712. Springer (2022)

\bibitem{X-Decoder}
Zou, X., Dou, Z.Y., Yang, J., Gan, Z., Li, L., Li, C., Dai, X., Behl, H., Wang, J., Yuan, L., et~al.: Generalized decoding for pixel, image, and language. In: CVPR. pp. 15116--15127 (2023)

\bibitem{seem}
Zou, X., Yang, J., Zhang, H., Li, F., Li, L., Gao, J., Lee, Y.J.: Segment everything everywhere all at once. arXiv preprint arXiv:2304.06718  (2023)

\end{thebibliography}
